\documentclass[runningheads]{llncs}

\usepackage{eccv}

\usepackage{eccvabbrv}

\usepackage{graphicx}
\usepackage{booktabs}

\usepackage[accsupp]{axessibility}  

\usepackage{amsmath}
\usepackage{algorithm}
\usepackage{algorithmic}

\usepackage{wrapfig}

\usepackage{multirow}
\usepackage{subcaption}
\usepackage{colortbl}

\definecolor{hight_light}{RGB}{224, 241, 239}
\definecolor{darkgray}{gray}{0.4}
\definecolor{mygreen}{RGB}{0,128,0}

\definecolor{seen_back}{RGB}{224, 241, 239}
\definecolor{unseen_back}{RGB}{243, 231, 213}
\definecolor{overall_back}{RGB}{220, 245, 220}

\usepackage[pagebackref,breaklinks,colorlinks,citecolor=eccvblue]{hyperref}

\usepackage{orcidlink}

\begin{document}

\title{SCORE: SubDistribution-aware Collaborative Knowledge Reinforcing for Cloth-Hybrid Lifelong Person Re-Identification} 


\author{Kunlun Xu\inst{1}\orcidlink{0000-0002-1706-4102} \and
Liangyu Ma\inst{1}\orcidlink{0009-0006-3423-1933} \and
 Jiangmeng Li\inst{2}\orcidlink{0000-0002-3376-1522}
 \and
Xin Tong\inst{3}\orcidlink{0000-0001-7424-0726}
 \and
 Xiaode Liu\inst{3}\orcidlink{0000-0003-3067-4543}
 \and
Yufei Guo\inst{3*}\orcidlink{0000-0002-4920-0965}
 \and
 Jiahuan Zhou\inst{1*}\orcidlink{0000-0002-3301-747X}
 }

\authorrunning{Kunlun Xu et al.}


\institute{Wangxuan Institute of Computer Technology, Peking University, Beijing, China 
\email{\{xkl,2300012949\}@stu.pku.edu.cn},
\email{jiahuanzhou@pku.edu.cn}
\and
University of Chinese Academy of Sciences, Beijing, China
\email{jiangmeng2019@iscas.ac.cn}
\and
Intelligent Science \& Technology Academy of CASIC, Beijing 100041, China\\
\email{\{xin\_tong,lxde,yfguo\}@pku.edu.cn}
}
\maketitle
\def\thefootnote{*}\footnotetext{Corresponding authors: Yufei Guo, Jiahuan Zhou}

\begin{abstract}
Lifelong Person Re-Identification (LReID) aims to train a unified person retrieval model from a non-stationary data stream. Existing LReID methods mainly focus on scenarios where the clothing of each person is consistent. Recently, the Cloth-Hybrid LReID (CH-LReID) where cloth-consistent and cloth-changing data alternately occur, has emerged as a more practical and challenging scenario. Due to the conflict between clothing-relevant and clothing-irrelevant knowledge, the well-known catastrophic forgetting problem is significantly exacerbated in this task. To address this issue, we propose a \textbf{S}ubDistribution-aware \textbf{CO}llaborative Knowledge \textbf{RE}inforcing (SCORE) framework, where our key idea is explicitly modeling the intra-identity diversity to continually consolidate distinct cloth-consistent and cloth-changing knowledge. Specifically, an Adaptive SubDistribution Modeling mechanism is developed, where a set of distributional subprototypes is assigned to each identity to capture the intra-identity diversity, improving the compatibility between cloth-consistent and cloth-changing knowledge. Then, a  Distributional Knowledge Reinforcement scheme is introduced, where the knowledge of old distributional subprototypes is retained in the new ones by a collaborative aligning mechanism. Extensive experiments show that our SCORE achieves the state-of-the-art performance.
 Our code is available at
\href{https://github.com/zhoujiahuan1991/ECCV2026-SCORE}{https://github.com/zhoujiahuan1991/ECCV2026-SCORE}.
\end{abstract}

\begin{figure}[ht!]
    \centering
	\includegraphics[width=0.75\linewidth]{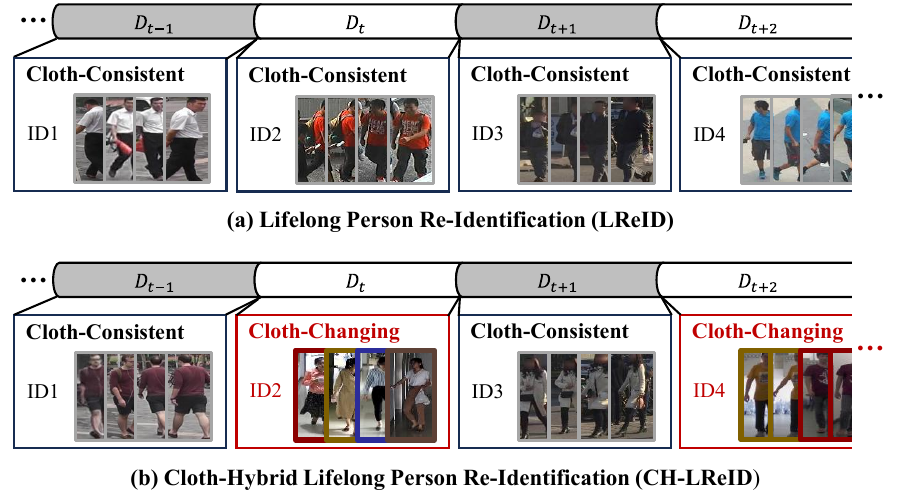}
        \caption{\label{fig:first} (a) Previous  LReID methods primarily focus on the cloth-consistent lifelong person re-identification scenario. (b) This paper addresses a more practical and challenging cloth-hybrid lifelong person re-identification problem where the cloth-consistent and cloth-changing training data occur alternatively.}
\end{figure}
\section{Introduction}
\label{sec:intro}

Person Re-Identification (ReID)~\cite{gongcross, shi2023dual} aims to match the same person across cameras. Traditional ReID methods focus on static scenarios with stationary cameras and environments~\cite{shi2024learning,yin2024robust,gong2022person,gong2024cross}. Recently, Lifelong Person Re-Identification (LReID)~\cite{wu2021generalising,pu2021lifelong,xu2025self,zhou2025distribution} has attracted increasing research interest by considering that non-stationary data arrives continually, where the catastrophic forgetting of past knowledge is the key challenge. However, existing LReID methods primarily consider that the clothing of each person is consistent~\cite{yu2023lifelong,huang2022lifelong,xu2024lstkc,cui2024continual,xu2025dask}, as shown in Fig.~\ref{fig:first} (a). Since lifelong learning usually involves long-term data collection, changes in people’s clothing inevitably occur from time to time, leading to a practical Cloth-Hybrid LReID (CH-LReID)~\cite{cui2025dkc}, as shown in Fig.~\ref{fig:first} (b).

\begin{figure}[ht]
    \centering
	\includegraphics[width=0.8\linewidth]{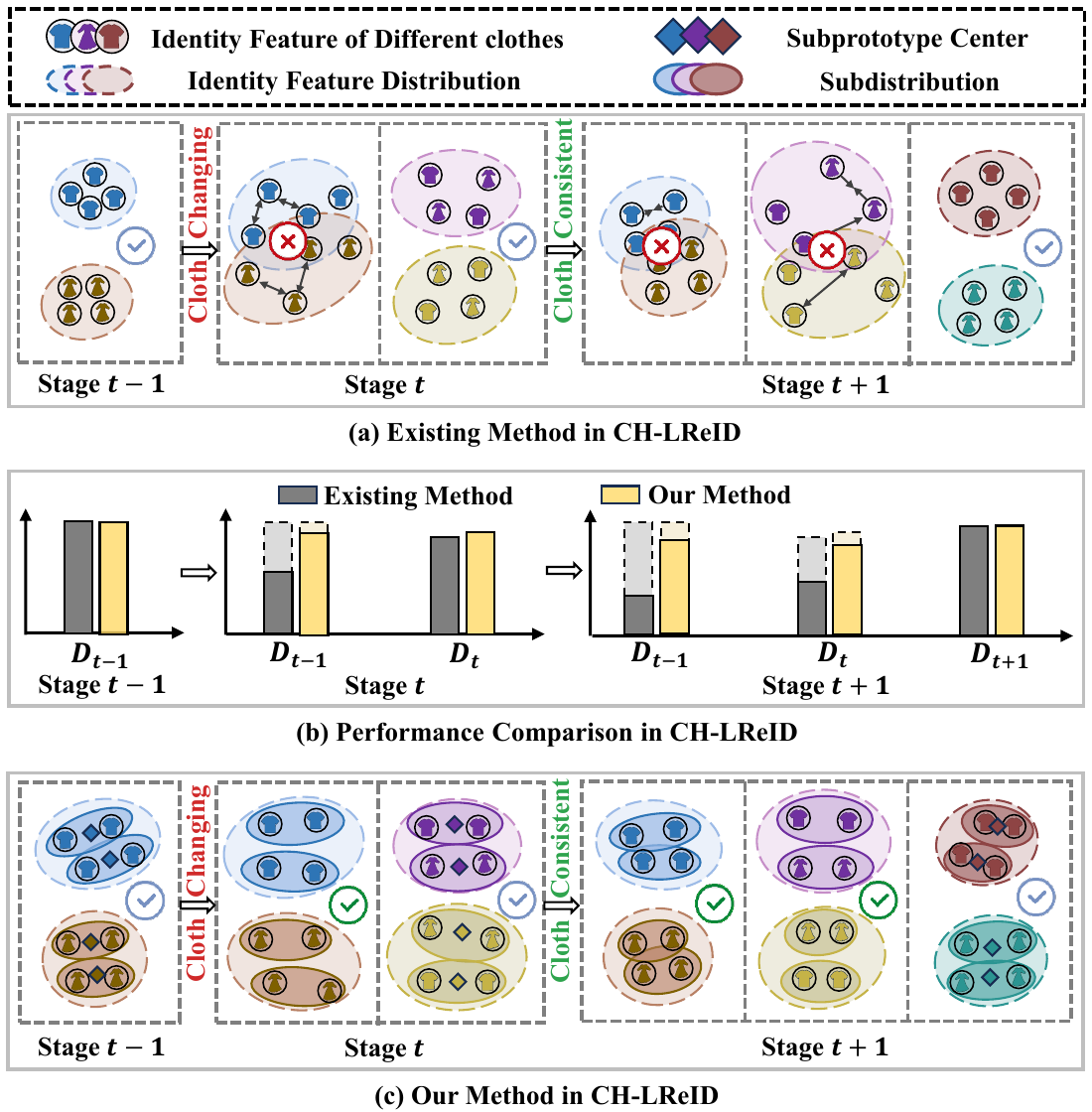}
        \caption{\label{fig:first-2} (a) Due to the optimization conflict between cloth-consistent and cloth-changing scenarios, the catastrophic forgetting problem emerges as a critical challenge in CH-LReID. (b)(c) We propose to explicitly model the intra-identity distribution diversity, improving the compatibility between the cloth-consistent and cloth-changing knowledge. Thereby, it significantly addresses the catastrophic forgetting problem .}
\end{figure}

In CH-LReID, the catastrophic forgetting problem is further exacerbated due to knowledge conflicts between training stages. As shown in Fig.~\ref{fig:first-2} (a), at stage $t-1$, the model learns from a cloth-consistent dataset, in which clothing cues help to identify people. However, when cloth-changing data is introduced at stage $t$, the inter-cloth matching constraints cause the model to forget the previously learned cloth-consistent knowledge. Later, at stage $t+1$, the model is trained again on cloth-consistent data, causing the overwriting of cloth-changing knowledge. In addition, the model’s performance on earlier domains remains limited because some old knowledge cannot be reproduced due to domain gaps across stages. This ongoing vanishment of cloth-consistent and cloth-changing knowledge results in severe performance degradation across all past domains.

Although Cloth-Changing ReID methods~\cite{gu2022clothes,han2023clothing,huang2021clothing} have been proposed in recent years, the issues in Fig.~\ref{fig:first-2} (a) persist when they are combined with LReID approaches. This is because these methods focus only on improving cloth-changing knowledge acquisition, without addressing the inherent knowledge conflict problem. Moreover, most of them rely on auxiliary information such as clothing labels, gait features, or body shape contours during training. However, obtaining these auxiliary data requires substantial manual effort, which limits their practicality in the lifelong learning scenario, which typically requires efficiency.

To address these limitations,  as shown in Fig.~\ref{fig:first-2} (c), we propose to explicitly model the intra-identity distribution diversity, thereby enabling the model to mine and store the cloth-consistent and cloth-changing knowledge simultaneously. As a consequence, inter-stage knowledge conflict can be effectively alleviated, making the catastrophic forgetting problem largely mitigated (as shown in Fig.~\ref{fig:first-2} (b)). 


To achieve this, we introduce a \textbf{S}ubDistribution-aware \textbf{CO}llaborative Knowledge \textbf{RE}inforcing (SCORE) framework. Firstly, an Adaptive SubDistribution Modeling (ASD) mechanism is developed, where the learnable distributional subprototypes are introduced to model the intra-identity subdistributions. Then, three optimization paradigms, Prototype Separating, Instance Assembling, and Distribution Purifying, are introduced to guide the prototypes to capture inter-identity subdistributions.
To mitigate the forgetting of abundant knowledge embedded in subdistributions from historical stages, a Distributional Knowledge Reinforcement (DKR) scheme is introduced to guide the new data distribution modeling by exploiting historical subdistributions, where a novel subprototype-level knowledge transfer loss is proposed to achieve interactive subdistribution modeling. To better retain the knowledge within the cross-instance representational relations, an Instance-oriented Structural Knowledge Retention (ISK) module is developed to further improve the overall anti-forgetting capacity. Extensive experimental results on the CH-LReID benchmark demonstrate that our approach outperforms existing approaches by large margins. In summary, our contributions are threefold:

(1) To address the cyclic knowledge conflict and amplified catastrophic forgetting in CH-LReID, we propose a novel approach SCORE that explicitly models the intra-identity variations to enhance the compatibility of the cloth-consistent and cloth-changing knowledge.

(2) An Adaptive SubDistribution Modeling mechanism is designed to automatically capture intra-identity subdistributions without relying on auxiliary data. Besides, the Distributional Knowledge Reinforcement scheme is developed to utilize historical subdistributional knowledge to guide new subdistribution learning, effectively achieving knowledge integration.

(3) Extensive experiments demonstrate the superiority of our proposed method in jointly consolidating the cloth-consistent and cloth-changing knowledge compared to the state-of-the-art approaches.

\section{Related Work}
\subsection{Lifelong Person Re-identification}
Lifelong Person Re-Identification (LReID) aims to learn from non-stationary data~\cite{pu2021lifelong,shi2024multi,li2024exemplar}, where catastrophic forgetting~\cite{li2024progressive,sun2022patch,xu2026vision} is the primary challenge. Existing LReID approaches can be divided into two streams. Data replay-based methods~\cite{wu2021generalising,yu2023lifelong} mitigate the forgetting of old knowledge by revisiting stored historical data during the learning of new domains. Although this method has achieved notable progress, growing concerns over data privacy~\cite{pu2022meta,yang2023handling} have drawn increasing attention to non-exemplar-based approaches. Knowledge distillation-based methods~\cite{xu2024mitigate,pu2022meta,sun2022patch} reduce the discrepancy between old and new domains by enforcing output consistency. Some methods, \textit{e.g.}, DKP~\cite{xu2024distribution}, introduce identity prototypes to reserve historical information. The above LReID methods typically focus on cloth-consistent scenarios where the cross-domain knowledge shares similar semantic clues~\cite{xu2024lstkc,yang2023handling,pu2021lifelong}. However, when applied to cloth-hybrid settings where the cloth-consistent and cloth-changing data show up alternately, the knowledge conflict caused by semantic clue change across domains would further intensify the catastrophic forgetting issue~\cite{xu2024lstkc}.

\subsection{Cloth-changing Person Re-identification}
Cloth-changing ReID is a challenging task where the diversity in clothing styles hinders the learning of robust knowledge.
To address this issue, some methods focused on disentangling clothing features during inference. 
Specifically, ~\cite{yang2023good, cui2023dcr} leverage clothing labels to force the model to rely less on clothing information. Besides, ~\cite{guo2023semantic,li2024disentangling} employ a semantic segmentation model to mask clothing information and introduce a consistency constraint between the original image and the clothing-removed image during training. 
Alternatively, other methods utilize information from additional modalities to learn clothes-irrelevant features, \textit{e.g.}, gait information~\cite{jin2022cloth} and silhouette information~\cite{yang2019person}.
However, their reliance on auxiliary information limits their application in practice. Recently, some methods have made notable advancements without introducing auxiliary information. 
For example,  methods~\cite{han2023clothing,liu2024cloth} employ data augmentation to diversify the colors and textures of clothing in the dataset, thereby reducing the association between identity and specific clothing. 
However, when faced with datasets that continuously come in a non-stationary manner, these methods suffer from catastrophic forgetting, where previously learned knowledge is overwritten, leading to degraded overall performance on seen domains.

\subsection{Distribution Learning}
Existing distribution learning methods primarily focus on modeling data uncertainty to address out-of-distribution (OOD) data ~\cite{yu2019robust,zheng2021rectifying,lu2023uncertainty,li2022uncertainty}. 
In the LReID task, DKP~\cite{xu2024distribution} performs instance-level modeling for each sample and aggregates them into prototype distributions, improving the model’s ability to acquire new knowledge. In this paper, targeting the cloth-hybrid scenario, we model multiple subdistributions within each identity to capture diverse discriminative knowledge adaptively, where learnable subprototypes are introduced to model the subdistributions. By collaboratively utilizing the different subdistributions learned from both cloth-consistent and cloth-changing data across stages, we aim to reduce knowledge conflicts and alleviate catastrophic forgetting.

\section{The Proposed Method}
\subsection{Preliminary}
In CH-LReID, a stream of $T$ cloth-hybrid training datasets $\mathcal{D}=\{D_t\}_{t=1}^{T}$ is provided sequentially. Specifically, given a dataset $D_t$ containing $N_t$ identities, the included data may be either cloth-consistent or cloth-changing, but the specific type is unknown in advance. Besides, for each identity, it is also not known beforehand whether they appear in multiple outfits. Furthermore, previously seen $t-1$ datasets are not accessible at the training stage $t$. 

\begin{figure*}[tb]
    \centering
	\includegraphics[width=1.0\linewidth]{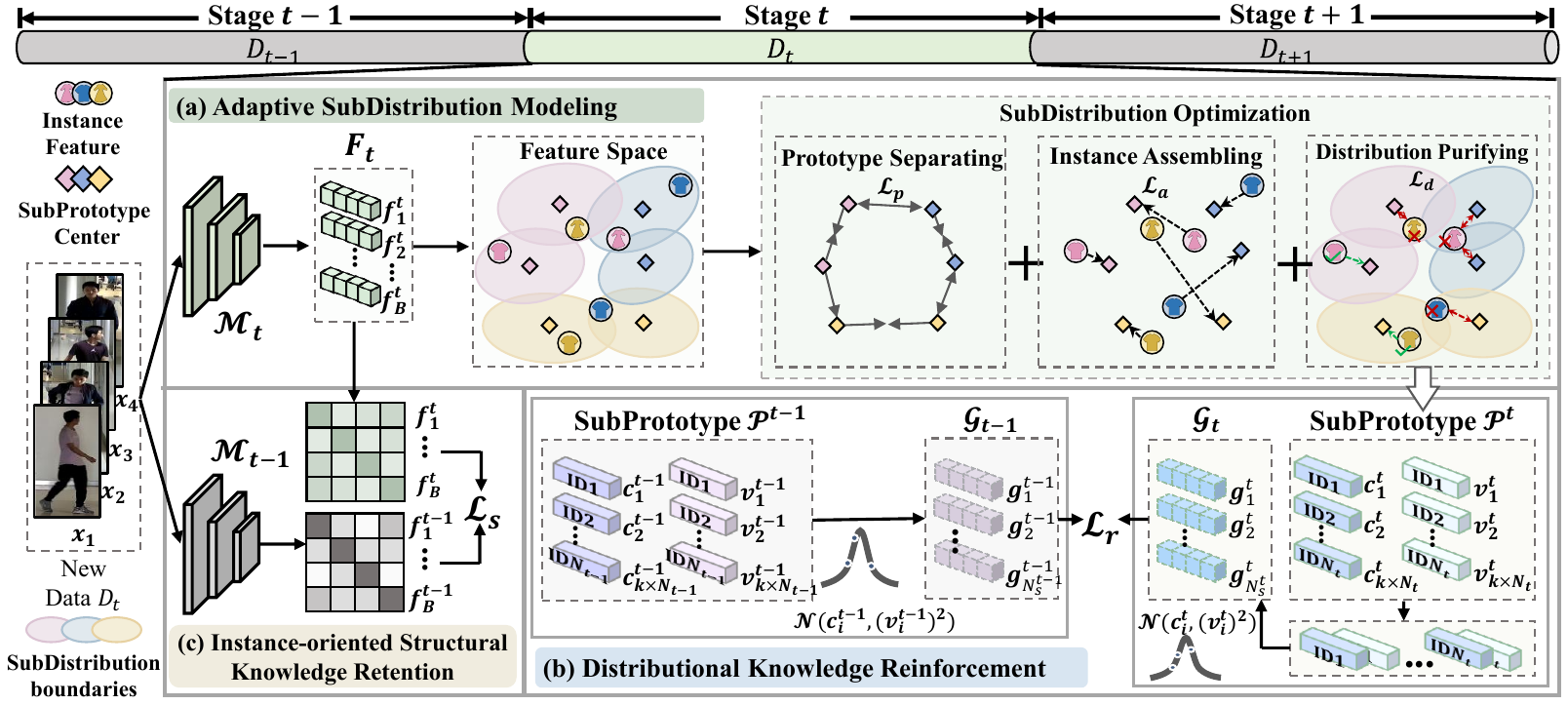} 
        \caption{The overview of our SCORE approach. (a) Adaptive SubDistribution Modeling (ASD) captures intra-identity diversity and encodes it into distributional subprototypes. (b) Distributional Knowledge Reinforcement (DKR) performs distributional alignment to guide new distribution learning with historical distribution. (c) Instance-oriented Structural Knowledge Retention (ISK)  integrates cross-instance relational knowledge across domains to improve the old knowledge consolidation. }
          \label{fig:framework} 
\end{figure*}

\subsection{Overview}
As shown in Figure.~\ref{fig:framework}, at the $t$-th training stage, given the new dataset $D_t$, we introduce $k$ learnable subprototypes in the feature space for each identity, denoted as $\mathcal{P}_t=\{(p^t_i,{y}^t_i)\}_{i=1}^{k \times N_t}$. Specifically, $p^{t}_{i}$ is a multi-dimensional Gaussian distribution, and ${y}^t_i$ is its corresponding identity label. The center of $p^{t}_{i}$ is denoted as $c^{t}_{i}\in \mathbb{R}^d$, and ${v^{t}_{i}}\in \mathbb{R}^d$ indicates the standard deviation vector derived from a diagonal covariance matrix $\Sigma^t_i \in \mathbb{R}^{d\times d}$ of $p^{t}_{i}$. Besides, we define $\mathcal{C}^t= \{c^t_i\}_{i=1}^{k \times N_t} $ and $\mathcal{V}^t = \{{v}^t_i\}_{i=1}^{k \times N_t}$ as the overall center set and standard deviation set at stage $t$, respectively.

The Adaptive SubDistribution Modeling (ASD) module aims to capture the intra-identity subdistributions via an adaptive distributional subprototype optimization mechanism. Besides, the Distributional Knowledge Reinforcement (DKR) module utilizes the preserved historical distributional information to guide the new subdistribution learning. Furthermore, the Instance-oriented Structural Knowledge Retention (ISK) module integrates fine-grained cross-instance relational knowledge from the previous stage into the new model to enhance the historical knowledge consolidation capacity.

\subsection{Adaptive SubDistribution Modeling}

Given a batch of input data ${ \{(x_i, y_i)\}}_{i=1}^B \in {D}_t$, where $B$ denotes the batch size, suppose there are $n$ distinct identities in this batch, and let the corresponding identity label set be $\mathcal{Y}={ \{y_i\}}_{i=1}^n$. The model backbone $\boldsymbol{M}_t$ extracts visual features from the inputs, producing a set of feature representations $\mathcal{F}=\{f_i\}_{i=1}^{B}$. To enhance the discriminative capability of the model, we employ the cross-entropy loss $L_{ce}$~\cite{xu2024lstkc} and the triplet loss $L_{tri}$ ~\cite{sun2022patch,xu2024mitigate} as the base supervision:
\begin{equation}
\left\{
\small
\begin{aligned}&
\mathcal{L}_{ce-s} = -\sum_{i=1}^{B} y_i \log\boldsymbol{\rho}(\textbf{W}_tf_{i})\\
&\mathcal{L}_{tri-s} = \log\left(1 + \exp\left(\left\| \tilde{f}_a - \tilde{f}_p \right\|_2^2 - \left\| \tilde{f}_a - \tilde{f}_n \right\|_2^2\right)\right)\\
\end{aligned}\right.\\
\label{eq:L_s_details},
\end{equation}
where $\textbf{W}_t$ denotes the linear projection parameter to map the features to logits, $\boldsymbol{\rho}$ represents the softmax function, $\tilde{f}$ is the $L_2$-normalized version of the feature, and $⟨a,p,n⟩$ is a triplet set. Therefore, the basic loss function $\mathcal{L}_{base}$ in LReID is formulated as:
\begin{equation}
\mathcal{L}_{base} = \mathcal{L}_{ce-s} + \mathcal{L}_{tri-s} 
    \label{eq:L_s}.
\end{equation}

In the following, we introduce the mechanism of subdistribution Optimization:

\textbf{Prototype Separating:} Prototype Separating aims to ensure the clustering of subprototypes of the same identity and separate the subprototypes of different identities. Specifically, for each input batch, we retrieve the corresponding prototypes based on $\mathcal{Y}$, denoted as  $\mathcal{P}^*=\{(p^t_i,{y}^t_i)\}_{i=1}^{k \times n}$, where ${y}^t_i \in \mathcal{Y}$. To ensure the coherence and the distinction among prototypes of the same and different identities respectively, we apply the cross-entropy loss and the triplet loss to $\mathcal{P}^*$, obtaining a prototype separating loss:
\begin{equation}
\mathcal{L}_{p} = \mathcal{L}_{ce-p} + \mathcal{L}_{tri-p} 
    \label{eq:L_proto},
\end{equation}

\textbf{Instance Assembling:} Instance Assembling aims to assign each instance to a subprototype and encourage the instance feature to approach the corresponding prototype. Since different outfits of the same identity may vary significantly in clothing style, color, and appearance cues, the knowledge encoded by each outfit can differ substantially. To capture this intra-identity diversity, each subprototype is modeled as a distribution, enabling different subprototypes to represent varying degrees of intra-identity variability. To measure how well an instance feature matches each subprototype, we compute the maximum likelihood of the instance feature under every subprototype. The resulting likelihood matrix $\mathcal{M} \in \mathbb{R}^{B \times kN_t}$ is formulated as:
\begin{equation}
\small
\mathcal{M}_{i,j}  = \frac{1}{(2\pi)^{d/2} |\Sigma^t_j|^{1/2}} \exp\left( -\frac{1}{2} (f_i - c^t_j)^T {(\Sigma^t_j)}^{-1} (f_i - c^t_j) \right)    
\label{}.
\end{equation}
where $i \in [1, B]$ and $j \in [1, kN_t]$.

Then, we pull each instance feature toward the most similar subprototype of the same identity. Accordingly, the instance assembling loss $\mathcal{L}_a$ is formulated as:
\begin{equation}
\mathcal{L}_a = \frac{1}{B} \sum_{i=1}^{B} (f_i - c_i^*)^2
\label{eq:L_r},
\end{equation}
where $c_i^*$ denotes the center of the most similar subprototype that shares the same identity label with $f_i$. 

\textbf{Distribution Purifying:} Distribution Purifying aims to ensure that each subdistribution exclusively models instances belonging to its corresponding identity. To achieve this, we first obtain the likelihood of the most compatible subprototypes from the likelihood matrix $\mathcal{M}$ for each instance, denoted as $\mathbf{m} \in \mathbb{R}^B$. A corresponding correctness indicator $\mathbf{z} \in \{0,1\}^B $ is also obtained which reflects whether the matched subprototype shares the same identity label:
\begin{equation}
\left\{
\begin{aligned}&
\mathbf{m}_i = \max_j \mathcal{M}_{i,j}\\
&\mathbf{z}_i = \mathbb{I}(y_i = y^t_{\arg\max_j \mathcal{M}_{i,j}})
\\
\end{aligned}\right.\\
\label{eq:ind},
\end{equation}
where $\mathbb{I}$ is an indicator function. Then, we design a distribution purifying loss $L_d$ which is computed as follows:
\begin{equation}
\small
\mathcal{L}_{d} = -\frac{1}{B} \sum_{i=1}^{B} \left[ \mathbf{z}_i \log(\boldsymbol{\rho}(\mathbf{m}_i)) + (1 - \mathbf{z}_i) \log(1 - \boldsymbol{\rho}(\mathbf{m}_i)) \right]
\label{eq:L_d}.
\end{equation}

\textbf{Overall SubDistribution Optimization Loss:} Prototype Separating, Instance Assembling, and Distribution Purifying mechanisms are complementary to each other since they perform inter-prototype, instance-centric, and prototype-centric SubDistribution evolution, respectively. Therefore, we compute the overall subdistribution optimization Loss as follows: 
\begin{equation}
    \mathcal{L}_{sub} = \mathcal{L}_{p} + \mathcal{L}_{a} + \mathcal{L}_{d}
    \label{}.
\end{equation}
 By incorporating the above constraints, our method encodes intra-identity differentiated knowledge in diverse subprototypes. Note that even for the cloth-consistent dataset, the subdistributions are applicable since all the distributions could be approximated as Gaussian mixtures~\cite{mclachlan2014number}.

\subsection{Distributional Knowledge Reinforcement}
In this section, we propose a collaborative aligning mechanism, which leverages the accumulated discriminative knowledge from the previous stage to guide the new subdistribution learning, thereby improving the coexistence of cloth-consistent and cloth-changing knowledge.

Specifically, as shown in Figure.~\ref{fig:framework}, when training in the new $t$-th stage, we sample $s$ subprototype features $\mathcal{G}_{{t-1}}=\{g^{t-1}_i\}_{i=1}^{N^{t-1}_s}$ from all subprototypes in the  $\mathcal{P}_{t-1}$ based on the distribution $\mathcal{N}(c_i^{t-1}, ({v}_i^{t-1})^2)$, where $N^{t-1}_s=s \times k \times N_{t-1}$. It is worth noting that, compared to directly using the subprototype centers, sampling from subdistributions allows for better utilization of the abundant distributional knowledge learned by each subprototype. 

For the $\mathcal{P}_{t}$ that are currently adaptively learnable, we only select the subprototypes that can be correctly matched according to Equation.~\ref{eq:ind} and also sample $s$ subprototype features $\mathcal{G}_{t}=\{g^{t}_i\}_{i=1}^{N^{t}_s}$, where $N^{t}_s=s \times n_c$ and $n_c$ denotes the number of correctly matched subprototypes. This is because the correctly matched subprototypes reflect the information that can currently accurately describe the identity, whereas the incorrectly matched subprototypes indicate confusion and bias in the knowledge they represent.

We convert $\mathcal{G}_{t-1}$ and $\mathcal{G}_{t}$ into matrix forms  $\mathbf{G}_{t-1} \in \mathbb{R}^{N_s^{t-1}\times d}$ and $\mathbf{G}_{{t}} \in \mathbb{R}^{N_s^{t} \times d}$, respectively. Then, a knowledge shift matrix $\mathbf{K}_s \in \mathbb{R}^{N_s^{t} \times N_s^{t-1}}$ for the subprototypes is obtained by:

\begin{equation}
\mathbf{K}_s = \boldsymbol\rho\left(\mathbf{G}_{{t}} \mathbf{G}_{{t-1}}^\top/{\lambda_1} \right)
\label{},
\end{equation}
where the softmax function $\boldsymbol\rho$ is applied across each row, and $\lambda_1$ serves as a temperature parameter to adjust the sharpness of the matrix values. To reduce the knowledge conflict caused by cloth-consistent and cloth-changing data across different stages, we perform distillation on the knowledge shift matrix $\mathbf{K}_s$ to collaboratively leverage both new and old knowledge. Accordingly, the subprototype-level knowledge transfer loss is defined as follows:
\begin{equation}
\mathcal{L}_{{t}} = \mathcal{KL}\left(\boldsymbol \rho\left(\mathbf{K}_s \mathbf{K}_s^\top/{\lambda_2} \right) \parallel \boldsymbol\rho\left( \mathbf{G}_{{t}} \mathbf{G}_{{t}}^\top/{\lambda_2} \right) \right) ,
\end{equation}
where $\mathcal{KL}$ denotes the Kullback-Leibler (KL) divergence and $\lambda_2$ is another temperature parameter.



\subsection{Instance-oriented Structural Knowledge Retention }
The DKR module facilitates the collaborative use of overall knowledge from both the new and old stages from the perspective of distributions. However, it lacks the integration of cross-instance fine-grained knowledge. To address this, inspired by the existing works~\cite{xu2024lstkc}, we propose an Instance-oriented Structural Knowledge Retention module. First, we use the old backbone $ \phi_{t-1}(\cdot)$, which was already trained in the $t-1$ stage, to extract features from the current input batch, obtaining the representations in the old feature space denoted as $ \mathbf{F}_o \in \mathbb{R}^{B \times d}$. Then, to preserve fine-grained old knowledge, we maximize the structural consistency of the input between the new and old feature spaces as follows:

\begin{equation}
\mathcal{L}_{{s}} = \mathcal{KL}\left(\boldsymbol \rho\left(\mathbf{F}_o \mathbf{F}_o^\top/{\lambda_2} \right) \parallel \boldsymbol\rho\left( \mathbf{F} \mathbf{F}^\top/{\lambda_2} \right) \right) 
\label{L_{ikr}},
\end{equation}
where $\mathbf{F} \in \mathbb{R}^{B \times d}$ is the matrix form of $\mathcal{F}$. Note that $ \boldsymbol \rho\left(\mathbf{F}_o \mathbf{F}_o^\top/{\lambda_2} \right) $ and $ \boldsymbol \rho\left(\mathbf{F} \mathbf{F}^\top/{\lambda_2} \right) $ both reflect the cross-instance representational relations, preserving the structured knowledge in the feature space. By guiding the model to focus on this structural consistency, the consolidation of fine-grained knowledge is significantly improved.

\subsection{Training and Inference}
During training, the overall loss $\mathcal{L}$ for SCORE can be computed as follows:
\begin{equation}
\mathcal{L} = \mathcal{L}_{base} + \alpha\mathcal{L}_{sub} + \beta(\mathcal{L}_{{t}}+\mathcal{L}_{{s}})
\label{},
\end{equation}
where $\alpha$ and $\beta$ are hyperparameters. Since both $\mathcal{L}_t$ and  $\mathcal{L}_s$ involves old knowledge, they share the same weight for simplicity. 

After training at the $t$-th stage, we fuse the newly learned model $\boldsymbol{M}_{t}$ with the old model $\boldsymbol{M}_{t-1}$ as a post-processing step following~\cite{xu2024distribution}:
\begin{equation}
\boldsymbol{M}_{t} \leftarrow \frac{\boldsymbol{M}_{t} + \boldsymbol{M}_{t-1}}{2}
\label{L}.
\end{equation}

During the inference phase, only the backbone is exploited to extract image features for person retrieval.

\section{Experiments}

\subsection{Datasets and Evaluation Protocol}

Following previous works~\cite{cui2025dkc},we evaluate the effectiveness of our method on the CH-LReID benchmark~\cite{cui2025dkc}, which consists of three cloth-consistent datasets (Market-1501~\cite{zheng2015scalable}, MSMT17-V2~\cite{wei2018person} and CUHK03~\cite{li2014deepreid}), as well as two cloth-changing datasets (LTCC~\cite{qian2020long}and PRCC~\cite{yang2019person}). 
Following the previous works~\cite{cui2025dkc}, we conduct training with two distinct dataset orders: (Order-1) Market → LTCC → PRCC → MSMT17 → CUHK03 and (Order-2) MSMT17 → PRCC → Market → CUHK03 → LTCC.

Following previous works~\cite{xu2024distribution,pu2021lifelong,cui2025dkc}, we adopt Rank-1 accuracy (R@1) and mean Average Precision (mAP) as evaluation metrics. To comprehensively verify the model's performance,  both cloth-consistent and cloth-changing setting are evaluated.

\subsection{Implementation Details}
To ensure a fair comparison with existing methods ~\cite{cui2025dkc,xu2024distribution,xu2024lstkc}, we adopt ResNet50~\cite{he2016deep} as the backbone network. For both training orders, the first dataset is trained for 80 epochs, while each of the subsequent datasets is trained for 60 epochs. We use a mini-batch size of 128, constructed by sampling 32 identities with 4 images per identity. Each image is resized to 256×128 during training. The model is optimized using SGD with a learning rate of 0.008 and a weight decay of 0.0001. Additionally, the temperature parameters $\lambda_1$and $\lambda_2$ are set to 0.8. All experiments are conducted on a single NVIDIA 4090 GPU.

\begin{table*}[t]
   \centering
   \caption{Performance comparison on Training Order-1: Market → LTCC → PRCC → MSMT17 → CUHK03.}
    \resizebox{\textwidth}{!}{
  \setlength{\tabcolsep}{0.05mm}
\begin{tabular}{l|cccccccccc>{\columncolor{seen_back}}c>{\columncolor{seen_back}}c|cccc>{\columncolor{unseen_back}}c>{\columncolor{unseen_back}}c|>{\columncolor{overall_back}}c>{\columncolor{overall_back}}c}
\hline
\multirow{3}[0]{*}{Method} & \multicolumn{12}{c|}{Cloth-Consistent} & \multicolumn{6}{c|}{Cloth-Changing} & \multicolumn{2}{>{\columncolor{overall_back}}c}{\textbf{Overall}} \\
\cline{2-19}
 
& \multicolumn{2}{c}{Market} & \multicolumn{2}{c}{LTCC}
& \multicolumn{2}{c}{PRCC} & \multicolumn{2}{c}{MSMT17}
& \multicolumn{2}{c}{CUHK03}
& \multicolumn{2}{>{\columncolor{seen_back}}c|}{\textbf{Average}} 
& \multicolumn{2}{c}{LTCC}   
& \multicolumn{2}{c}{PRCC}
& \multicolumn{2}{>{\columncolor{unseen_back}}c|}{\textbf{Average}} 
& \multicolumn{2}{>{\columncolor{overall_back}}c}{\textbf{Average}} \\

& mAP   & R@1   & mAP   & R@1   & mAP   & R@1   & mAP   & R@1   & mAP   & R@1   & mAP   & R@1   & mAP   & R@1 & mAP   & R@1& mAP   & R@1& mAP   & R@1\\
\hline

JointTrain     & 64.1 & 82.5 & 42.6 & 62.1 & 94.6 & 98.7 & 18.4 & 40.8 & 44.4 & 46.4 & 52.8 & 66.1 & 10.1 & 23.0 & 32.7 & 33.8 & 21.4 & 28.4 & 43.8 & 55.3  \\
SFT   & 28.5 & 52.0 & 28.5 & 49.3 & 92.5 & 97.3 & 7.0 & 19.6 &44.0 & 45.6 & 40.1 & 52.8 & 6.9 & 15.3 & 21.8 & 21.2 & 14.4 & 18.3 & 32.7&42.9 \\

\hline

LwF\cite{li2017learning}  & 44.5 & 65.8 & 21.6 & 40.0 & 87.4 & 91.3 & 4.0 & 11.6 & 25.5 & 25.0 & 36.6 & 46.7 & 5.9 & 12.5 & 25.9 & 26.7 & 15.9 & 19.6  & 30.7 & 39.0\\ 
AKA\cite{pu2021lifelong}  & 48.0 & 69.5 & 25.4 & 45.1 & 88.1 & 93.3 & 4.2 & 12.0 & 31.2 & 31.2 & 39.4 & 50.2 & 6.5 & 12.8 & 26.5 & 26.7 & 16.5 & 19.8 & 32.8 &41.5  \\
PatchKD\cite{sun2022patch}   &\textbf{68.0} & \textbf{85.5} & 30.8 & 54.7 & 93.5 & 96.5 & 5.7 & 15.6 & 33.2 & 32.9 & 46.2 & 57.0 & 7.2 & 17.9 & 26.1 & 26.0 & 16.7 &22.0 &37.8&47.0 \\    

LSTKC\cite{xu2024lstkc}    & 39.9 & 63.4 & 39.6 & 65.4 & 95.9 & 98.9 & 11.5 & 29.2 & \textbf{48.1} &\textbf{ 50.1} & 47.0 & 61.4 & 8.3 & 19.4 & 24.0 & 22.9 & 16.2 & 21.2 &38.2 & 49.9 \\
USP\cite{yan2024unified} & \underline{66.3} & 74.9 & 37.2 & 49.6 & 88.8 & 92.6 & 6.6 & 14.6 & 38.7 & 33.9 & 47.5 & 53.1 & 7.8 & 17.3 & 25.7 & 23.6 & 16.8 & 20.5 &38.7 & 43.8 \\

DKP\cite{xu2024distribution} & 51.3 & 73.0 & 44.2 & 68.6& {98.4} & \underline{99.4} &13.5 & 31.6 & 39.5 & 40.6 & 49.4 & 62.6 & {9.3} & {21.4} & 35.5 & {34.8} & {22.4} & {28.1} &41.7 &52.8\\

DASK\cite{xu2025dask}  & 56.8 & 78.6 & 32.7 & \underline{72.4}& \underline{98.5} & \underline{99.4} &\textbf{25.5} & \textbf{52.8} & \underline{42.4} & 44.4 & {51.2} & \underline{69.5} & \textbf{11.3} & \textbf{26.5} & {33.6} & {31.6} & 22.5 & {29.0} & 43.0 & \underline{57.9}\\
DSIFLF${}^\dag$\cite{li2024disentangling} &47.5	&69.2	&42.1	&65.5	&97.0	&97.8	&11.0	&26.4	&39.7 &41.6	&47.5 &60.1 &8.9	&17.9	&34.0	&31.7	&21.5	&24.8 & 40.0 & 50.0	\\
DKC\cite{cui2025dkc} &57.9 &78.5 &\underline{49.2} &\underline{72.4} &98.6 &\textbf{99.5} &16.2 &36.7 &40.7 &42.4 &\underline{52.5} &65.9 &10.1 &25.5 &\textbf{36.9} &\underline{35.5} &\underline{23.5} &\underline{30.5} & \underline{44.2} & {55.8}\\
\hline
\textbf{SCORE\ } & 63.1	&\underline{82.9}	&\textbf{58.2}	&\textbf{78.4}	&\textbf{98.7}	&\textbf{99.5}	&\underline{23.5}	&\underline{50.2}&	44.0&	\underline{45.2}	&\textbf{57.5}&	\textbf{71.2}
	&\underline{11.0}	&\underline{25.8}	&\underline{36.2}	&\textbf{36.4}	&\textbf{23.6}	&\textbf{31.1} &\textbf{47.8} & \textbf{59.7}

\\
    \hline  
    \end{tabular}%
    }
    \raggedright
\label{tab:setting1}%
\end{table*}%

\subsection{Comparison with State-Of-The-Art methods}
We compare our proposed SCORE method against the current state-of-the-art CH-LReID method DKC~\cite{cui2025dkc}. Besides, we also compare with LReID methods, including LwF~\cite{li2017learning}, AKA~\cite{pu2021lifelong}, PatchKD~\cite{sun2022patch}, LSTKC \cite{xu2024lstkc}, USP~\cite{yan2024unified},  DKP~\cite{xu2024distribution} and DASK~\cite{xu2025dask}. In addition, we incorporate the state-of-the-art cloth-changing ReID method DSIFLF~\cite{li2024disentangling}, with the anti-forgetting strategy of LReID. The resulting model is named DSIFLF${}^\dag$. Moreover, we include both sequential finetuning and joint training settings. The best and second-best results are highlighted in \textbf{bold} and \underline{underline}, respectively.

\begin{table*}[t]
   \centering
   \caption{Performance comparison on Training Order-2: 
MSMT17 → PRCC → Market-1501 → CUHK03 → LTCC.}
 \resizebox{\textwidth}{!}{
  \setlength{\tabcolsep}{0.1mm}
\begin{tabular}{l|cccccccccc>{\columncolor{seen_back}}c>{\columncolor{seen_back}}c|cccc>{\columncolor{unseen_back}}c>{\columncolor{unseen_back}}c|>{\columncolor{overall_back}}c>{\columncolor{overall_back}}c}
\hline
\multirow{3}[0]{*}{Method}  & \multicolumn{12}{c|}{Cloth-Consistent} & \multicolumn{6}{c|}{Cloth-Changing} & \multicolumn{2}{>{\columncolor{overall_back}}c}{\textbf{Overall}} \\
\cline{2-19}
&  
 \multicolumn{2}{c}{MSMT17} & \multicolumn{2}{c}{PRCC}
& \multicolumn{2}{c}{Market} & \multicolumn{2}{c}{CUHK03}
& \multicolumn{2}{c}{LTCC}
& \multicolumn{2}{>{\columncolor{seen_back}}c|}{\textbf{Average}} 
& \multicolumn{2}{c}{PRCC}   
& \multicolumn{2}{c}{LTCC}
& \multicolumn{2}{>{\columncolor{unseen_back}}c|}{\textbf{Average}} 
& \multicolumn{2}{>{\columncolor{overall_back}}c}{\textbf{Average}} \\

& mAP   & R@1   & mAP   & R@1   & mAP   & R@1   & mAP   & R@1   & mAP   & R@1   & mAP   & R@1   & mAP   & R@1 & mAP   & R@1& mAP   & R@1& mAP   & R@1\\
\hline

JointTrain    & 18.4 & 40.8 & 94.6 & 98.7 & 64.1 & 82.5& 44.4 & 46.4& 42.6 & 62.1 & 52.8 & 66.1 & 32.7 & 33.8 & 10.1 & 23.0 & 21.4 & 28.4 & 43.8 & 55.3  \\
SFT     & 2.4 & 9.1 & 84.5 & 95.9 & 16.9& 37.2 & 8.2& 8.8 & 37.9 & 57.6 & 30.0 & 41.7 & 21.9 & 24.3 & 9.1 & 20.9 & 15.5 & 22.6 & 25.8 & 36.3  \\

\hline

LwF\cite{li2017learning}   &12.7 &28.8 &91.3 &95.5 &25.2 &49.0 &8.6 &8.0 &41.9 &63.3 &35.9 &48.9 &27.9 &28.3 &8.7 &18.4 &18.3 &23.4 & 30.9 & 41.6 \\ 
AKA\cite{pu2021lifelong}   &15.0 &32.4 &93.0 &97.0 &26.6 &51.0 &13.4 &11.6 &43.4 &64.0 &38.3 &51.2 &28.2 &27.7 &8.4 &15.8 &18.3 &21.8 & 32.6 & 42.8  \\
PatchKD\cite{sun2022patch} & \underline{22.6} &\textbf{47.9} &95.0 &98.5 &38.2 &64.2 &21.8 &23.1 &45.8 &70.0 &44.7 &60.7 &28.1 &27.6 &9.3 &21.7 &18.7 &24.7 & 37.3 & 50.4    \\    

LSTKC\cite{xu2024lstkc}   &6.9 &20.4 &92.8 &97.0 &39.8 &64.3 &25.2 &26.0 &51.1 &65.5 &43.2 &54.6 &27.3 &27.0 &10.9 &22.4 &19.1 &24.7 & 36.3 & 46.1  \\
USP\cite{yan2024unified}  &\textbf{22.7} &37.2 &92.5 &95.9 &37.8 &53.5 &17.3 &14.6 &\underline{63.5} &72.4 &46.8 &54.7 &28.7 &29.4 &11.9 &{26.0} &20.3 &27.7 & 39.2 & 47.0  \\

DKP\cite{xu2024distribution}& 11.8 &29.0 &{96.7} &{98.8} &46.0 &69.8 &{29.9} &{30.9} &55.8 &{75.8} &{48.0} &{60.9} &{34.1} &{34.6} &11.6 &24.2 &{22.9} &{29.4} & 40.8 & 51.9 \\
DASK\cite{xu2025dask} & 13.3 & 35.0 & \underline{98.2} & \underline{99.4}& {51.3} & \underline{74.3} &{26.1} & {25.5} & {35.3} & {72.4} & {44.8} & {61.3} & {34.8} & {31.9} & \underline{13.5} & \underline{27.8} & {24.1} & {29.8} & 38.9 & 52.3
 \\
DSIFLF${}^\dag$\cite{li2024disentangling}&8.6	&22.4	&{96.7}	&97.2	&{50.0}	&{71.4}	&25.8	&25.7	&57.4	&74.8	&47.7	&58.3	&32.8	&30.3 &{12.8}	&23.2	&22.8	&26.8 & 40.6 & 49.3 \\

DKC\cite{cui2025dkc}&15.4	&35.3	&98.1	&99.0	&\underline{51.8}	&72.7	&\underline{31.9}	&\underline{32.1}	&57.2 &\underline{77.9} &\underline{50.9} &\underline{63.4} &\textbf{38.2} &\textbf{37.9} &13.0 &27.0 &\textbf{25.6} &\underline{32.5} & \underline{43.7} & \underline{54.6} 
\\
\hline
\textbf{SCORE\ } & \underline{19.9}&	\underline{44.5}&	\textbf{98.9}&	\textbf{99.5}	&\textbf{55.3}&	\textbf{77.3}	&\textbf{34.5}&	\textbf{35.1}	&\textbf{65.0}	&\textbf{81.1}	&\textbf{54.7}	&\textbf{67.5} &\underline{36.5}	&\underline{35.4} &\textbf{14.4}	&\textbf{32.7}	&\underline{25.4}	&\textbf{34.0} & \textbf{46.4} & \textbf{57.9} \\

    \hline  
    \end{tabular}%
   }
 \raggedright \\
\label{tab:setting2}%
\end{table*}%

\textbf{Compared on the Cloth-Consistent data}:
As shown in Table.~\ref{tab:setting1} and Table.~\ref{tab:setting2}, our SCORE outperforms all existing methods under the Cloth-Consistent. Specifically, on Training Order-1, SCORE surpasses the state-of-the-art DASK and DKC with \textbf{5.0\%} and \textbf{1.7\%} Average mAP and R@1 improvements, respectively. Furthermore, on Training Order-2, SCORE outperforms the state-of-the-art DKC with \textbf{3.8\%/4.1\%} Average mAP/R@1 improvements. 
These results demonstrate the effectiveness of our method in the CH-LReID scenario, where it can distinguish individuals by effectively integrating both clothing-related and clothing-irrelevant knowledge learned across different training stages, rather than suppressing the use of clothing information. 

These advantages arise from our SubDistribution-aware Collaborative Knowledge Reinforcing mechanism, which effectively alleviates the conflict between cloth-changing and cloth-consistent knowledge, thereby improving the cross-domain knowledge accumulation capacity. In contrast, existing LReID works, \textit{e.g.}, DASK, suffer from performance limitations as they discard clothing information in the feature space for each identity, which inevitably hinders the effective use of valuable clothing-related cues. The existing  CH-LReID method, DKC, though considered to estimate the subdistributions, only exploited the statistical distribution and neglected the dynamic modeling of subdistributions. Thus, it suffered from limited representation and utilization of inter-identity diversity.

\textbf{Compared on the Cloth-Changing data}:
As shown in
Table.~\ref{tab:setting1}, SCORE outperforms state-of-the-art DKC by \textbf{0.1\%/0.6\%} under Average mAP/R1 on the Cloth-Changing data under Training-Order-1. Besides, under Training-Order-2, our SCORE surpasses DKC by \textbf{1.5\%} Average R@1 and obtains comparable performance on  Average mAP. This indicates that, for cloth-changing data, our approach effectively captures and preserves the identity-specific differentiated knowledge introduced by variations in clothing. By collaboratively leveraging diverse discriminative knowledge, it mitigates the conflict induced by data type switching.

\textbf{Compared on the Overall Performance}: 
As shown in Table.~\ref{tab:setting1} and Table.~\ref{tab:setting2}, our SCORE achieves state-of-the-art overall performance compared to the existing approaches. Specifically, SCORE outperforms the existing methods by a minimal improvement of \textbf{3.6\%/1.8\%} and \textbf{2.7\%/3.3\%} under Training Order-1 and Training-Order-2, respectively. These results arise from the proposed SubDistribution-aware Collaborative Knowledge Reinforcing paradigm that effectively improves the compatibility between the Cloth-Changing and Cloth-Consistent knowledge.

\begin{figure}[t]
    \centering
	\includegraphics[width=0.8\linewidth]{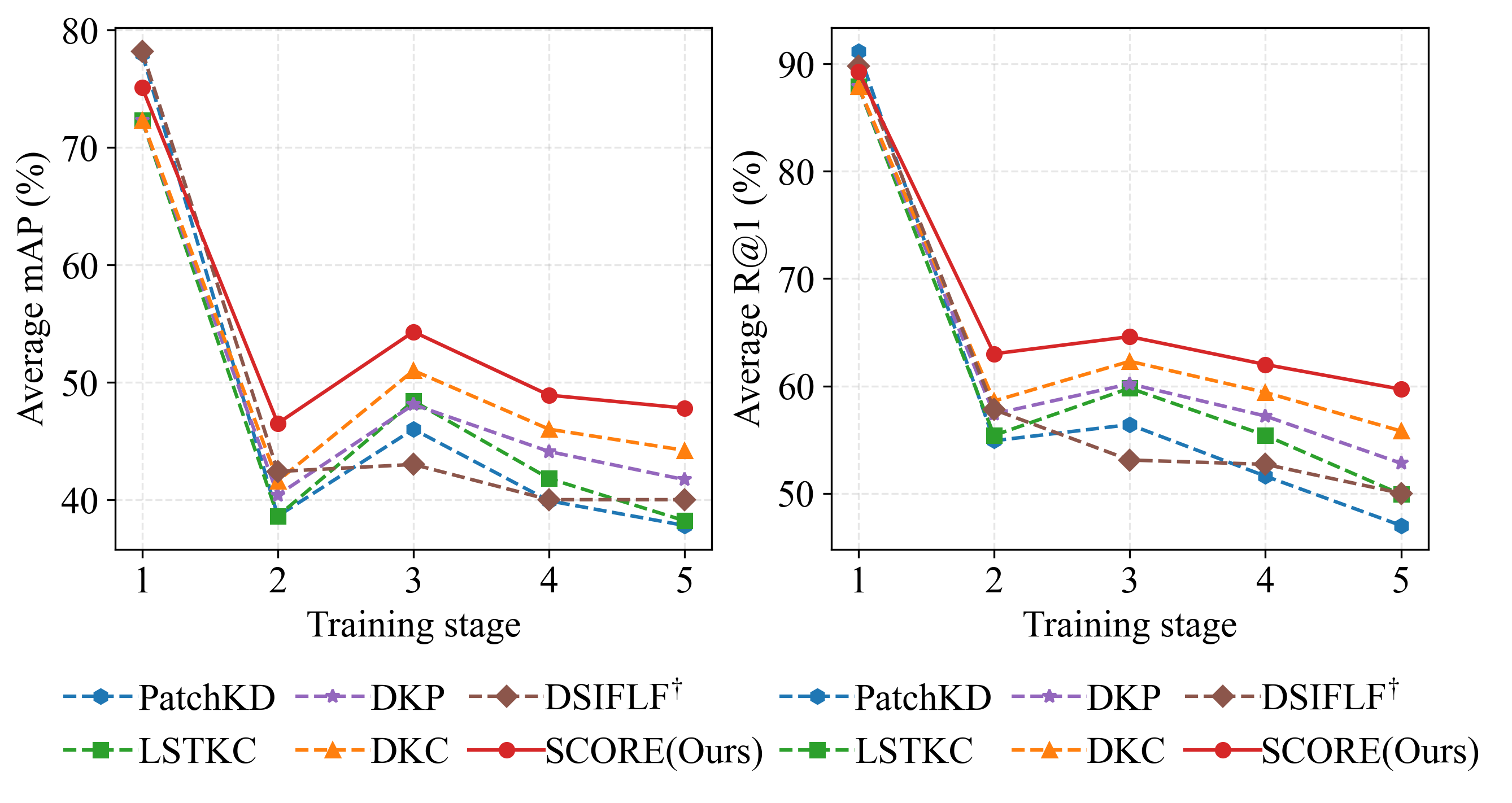}
        \caption{\label{fig:forggetting} Performance tendency at different training stages. }
\end{figure}

\textbf{Performance Tendency in Cloth-Hybrid Scenario}: 
Figure.\ref{fig:forggetting} illustrates the overall performance tendency of different methods on seen domains. The results show that our SCORE, although exhibiting comparable performance with existing methods, shows a consistent improvement over existing approaches. This highlights that our method is capable of consistent learning in cloth-hybrid scenarios, and achieves superior anti-forgetting performance by integrating diverse discriminative knowledge to alleviate knowledge conflicts.

\begin{table}[t!]
  \centering
  \caption{Ablation study of different components.}
  \setlength{\tabcolsep}{1.1mm}{
    \begin{tabular}{cccccccccc}
    \hline
    && && \multicolumn{2}{c}{Cloth-Consist. }& \multicolumn{2}{c}{Cloth-Chang.} \\
    \multirow{-1.9}[0]{*}{Base}& \multirow{-1.9}[0]{*}{ASD} & \multirow{-1.9}[0]{*}{ISK} & \multirow{-1.9}[0]{*}{DKR}   & mAP   & R@1 &mAP   & R@1\\ 
          \hline
    $\checkmark$  &             &               & &40.1 &52.8 &14.4 &18.3 \\
    $\checkmark$  &$\checkmark$ &                & &50.1	&63.9 &22.3	&27.6\\             
    $\checkmark$  &$\checkmark$ &$\checkmark$     &  & {52.8} & {66.3} & 23.0 & 29.6 \\
    $\checkmark$  &$\checkmark$ &$\checkmark$    &$\checkmark$ & \textbf{57.5} & \textbf{71.2} &\textbf{23.6} & \textbf{31.1} \\
    \hline
    \end{tabular}%
 }
  \label{tab:componment}%
\end{table}%


\subsection{Ablation Study}

\textbf{Ablations on model components.}
As shown in Table.~\ref{tab:componment}, we adopt Sequential Finetuning (SFT) as the baseline and incrementally incorporate the ASD, ISK, and DKR modules. 
When ASD is adopted, the model achieves \textbf{10.0\%/11.1\%} and \textbf{7.9\%/9.3\%} improvements in mAP/R@1 on Cloth-Consistent and Cloth-Changing data, respectively. This improvement is attributed to the SubDistribution modeling mechanism, which enhances the compatibility between Cloth-Consistent and Cloth-Changing knowledge, thereby mitigating the catastrophic forgetting issue.
Furthermore, as ISK is introduced, additional \textbf{2.7\%/2.4\%} and \textbf{0.7\%/2.0\%} improvements under mAP/R@1 metrics on Cloth-Consistent and Cloth-Changing data are achieved. This is because ISK transfers the inter-instance structure knowledge, which is complementary to the clothing knowledge. 
Finally, incorporating DKR yields additional gains of \textbf{4.7\%/4.9\%} and \textbf{0.6\%/1.5\%}, resulting in total improvements of \textbf{17.4\%/18.4\%} and \textbf{9.2\%/12.8\%}. These gains stem from explicitly leveraging historical distributions to guide new model learning, further alleviating conflicts between Cloth-Consistent and Cloth-Changing knowledge.

\begin{wrapfigure}{!t}{0.35\textwidth}
\vspace{-0.8cm}
\includegraphics[width=1\linewidth]{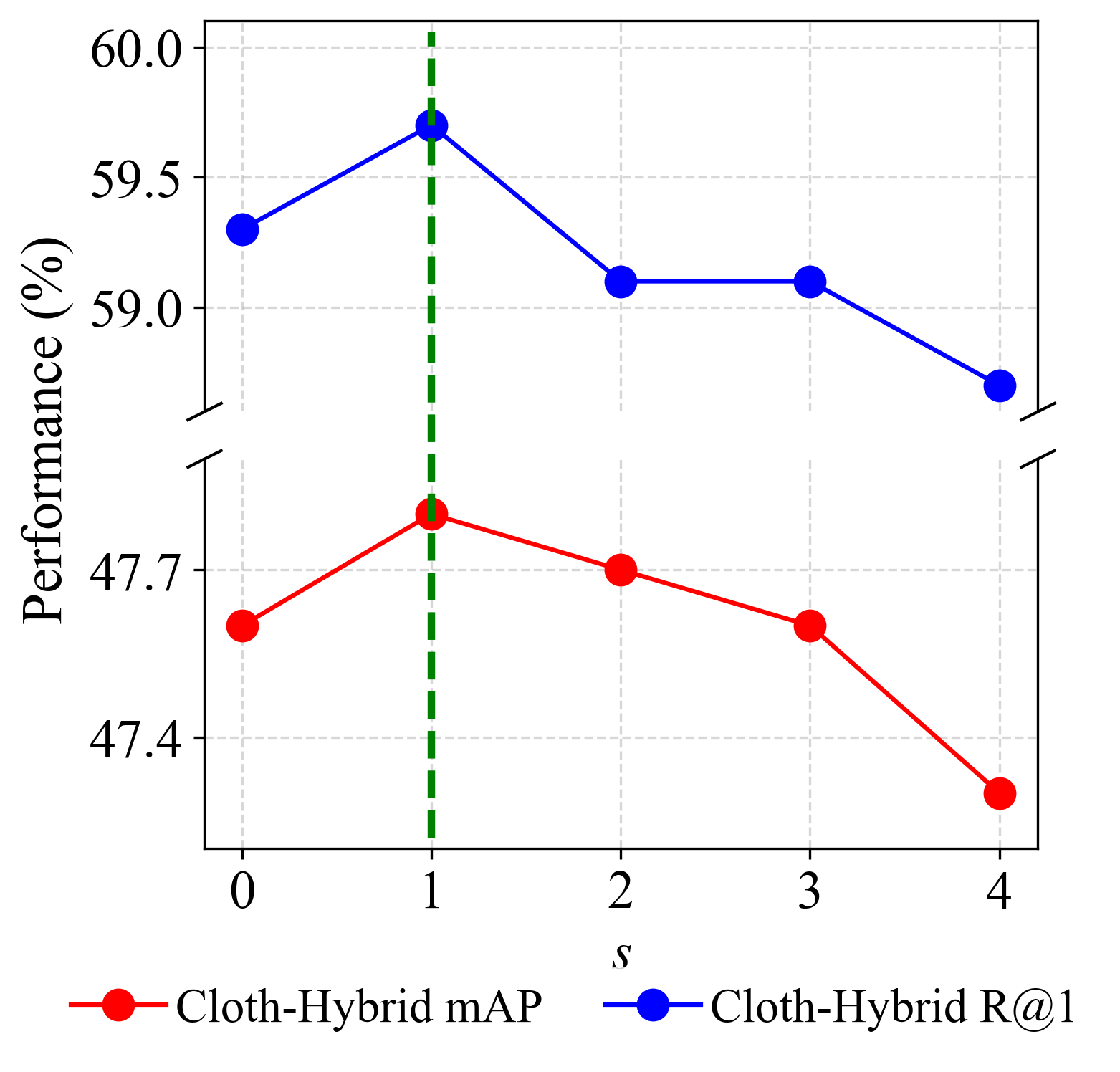}
\vspace{-0.5cm}
\caption{Ablation studies on feature sample number $s$. }
\label{fig:ablation1}
\vspace{-0.5cm}
\end{wrapfigure}

Note that the performance improvement of the Cloth-Changing data is less significant on the Cloth-Consistent data. This is because the Cloth-Changing scenario is inherently more challenging, making performance gains more difficult to achieve.

\begin{figure*}[t!]
		\centering
            \subfloat[\label{fig:alpha} \textbf{The weight of $\mathcal{L}_{sub}$}]{
                \includegraphics[scale=0.095]{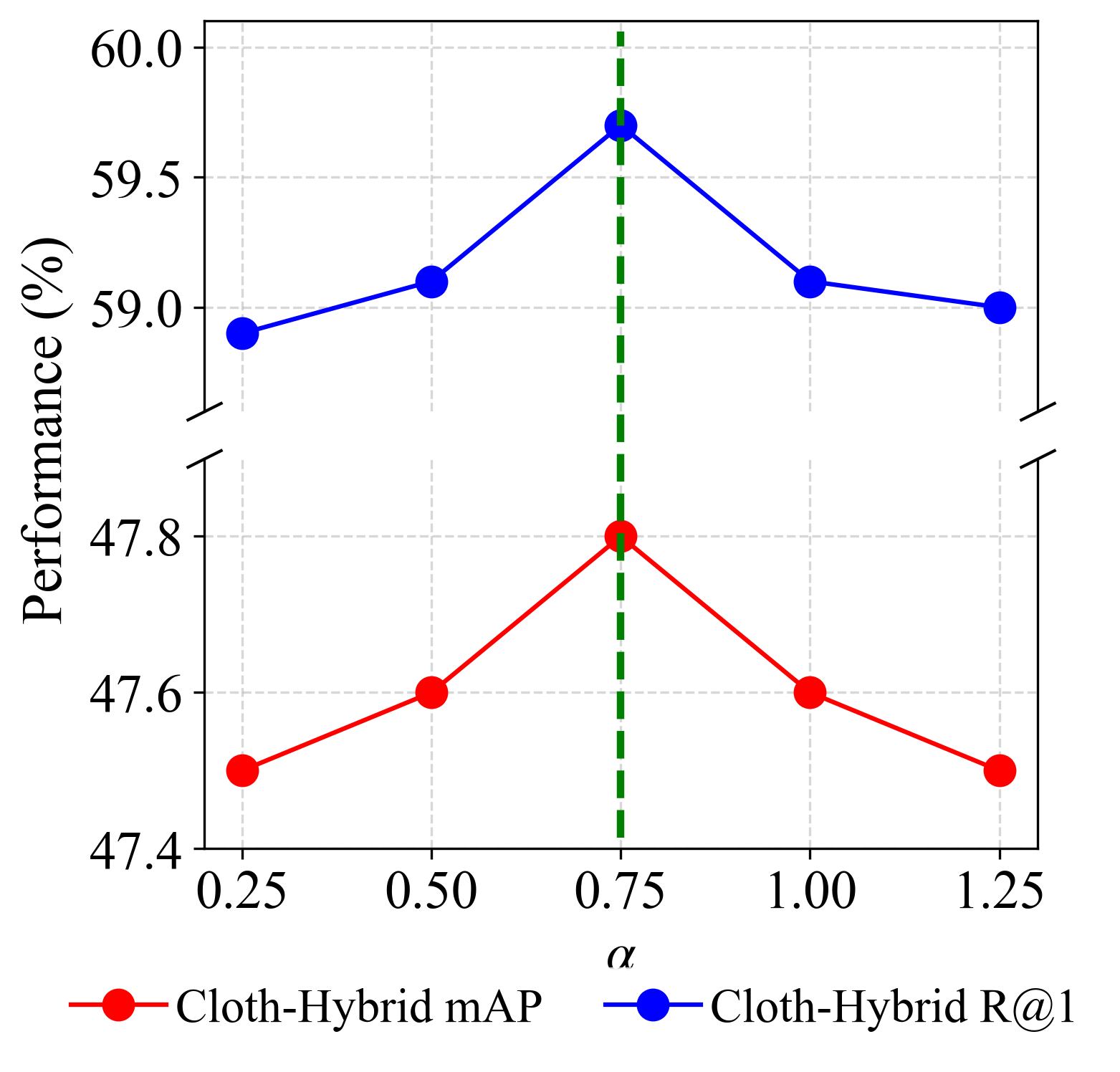}}
            \subfloat[\label{fig:beta}\textbf{The weight of $\mathcal{L}_{t}+\mathcal{L}_{s}$}]{
                \includegraphics[scale=0.095]{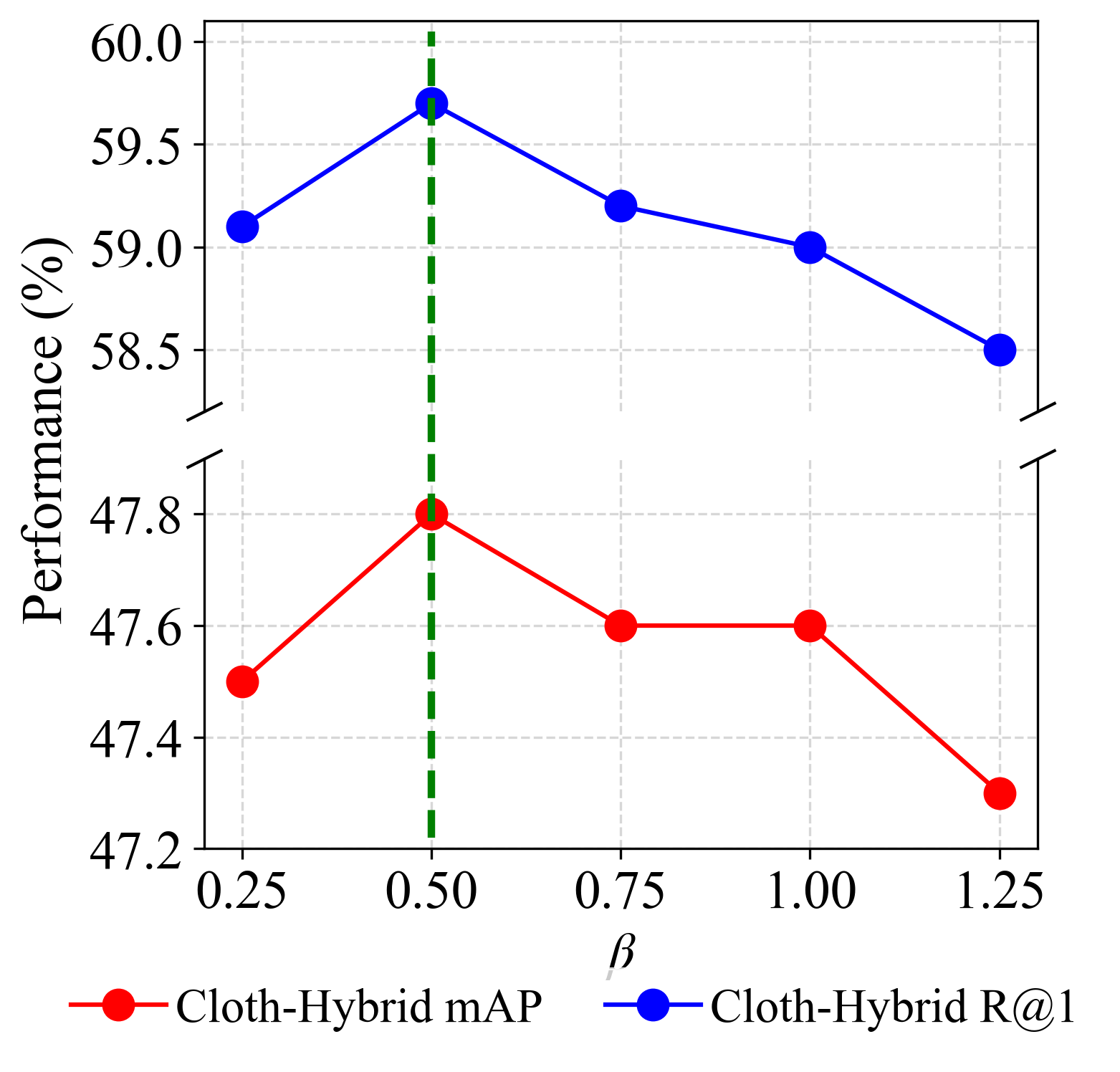}}
            \subfloat[\label{fig:proto-number} \textbf{SubDistribution per ID $k$}]{
                \includegraphics[scale=0.095]{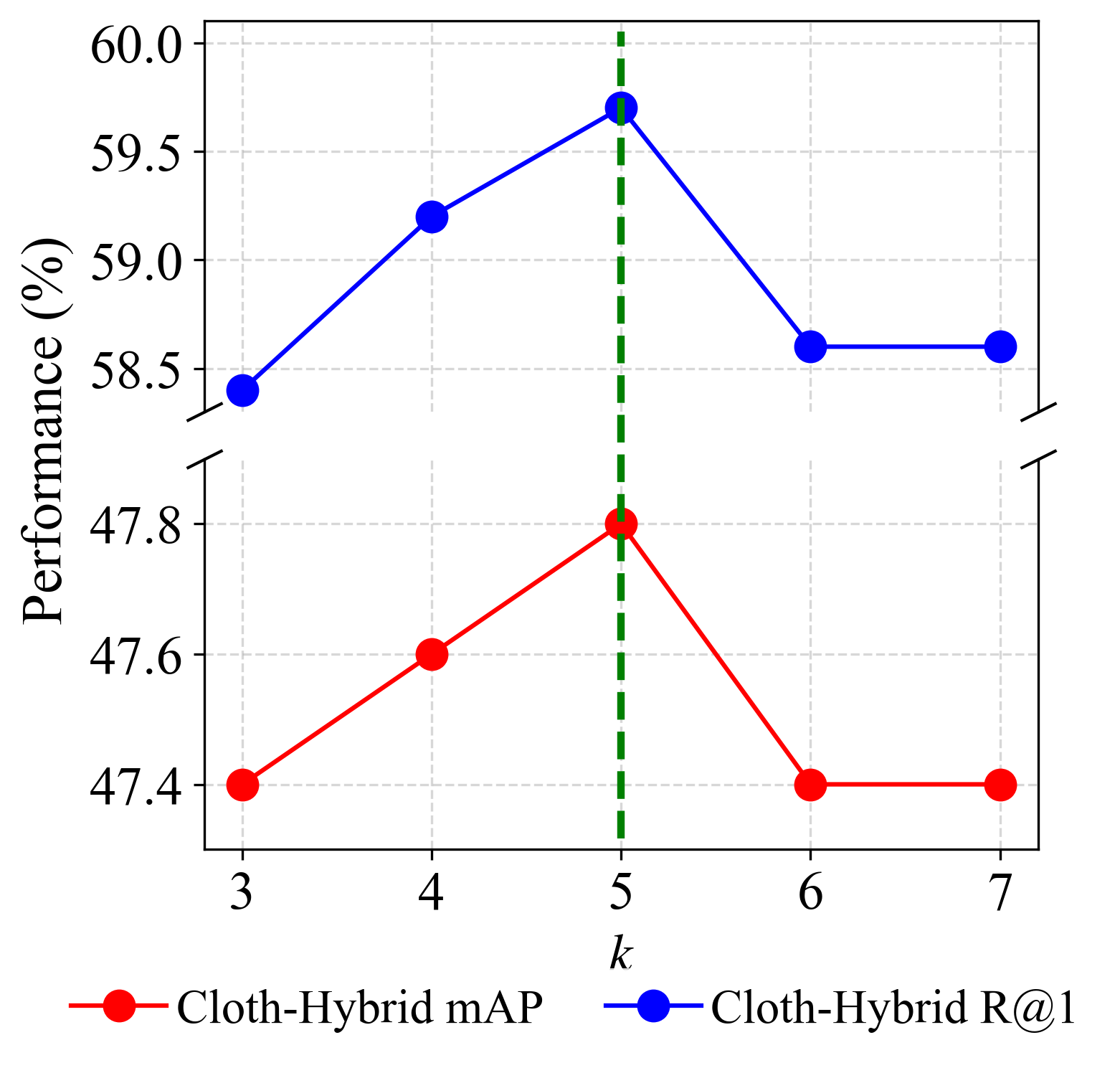}}
		\caption{Ablation studies on hyperparameters under training Order-1. The default values are highlighted by the dashed lines.
		}
		\label{fig:ablation} 
\end{figure*}

\textbf{Ablations on hyperparameters.}
We investigate the influence of the hyperparameters $\alpha, \beta, s, k$, and present the results in Figure.~\ref{fig:ablation} and Figure.~\ref{fig:ablation1}. $\alpha$ and $\beta$ are the loss weights to emphasize the subdistribution modeling and historical knowledge utilization, respectively. The proper $\alpha$ and $\beta$ can facilitate model learning, while too large values can lead to overfitting and limited robustness. Similarly, a proper predefined subdistribution number per class $k$ and instance sample number $s$ can promote model learning, while a too large $k$ and $s$ can cause optimization instability and overfitting.
In this paper, we empirically set $\alpha$, $\beta$, $k$, $s$ to 0.75, 0.50, 5 and 1 by default.


\begin{figure}[t]
    \centering
	\includegraphics[width=1\linewidth]{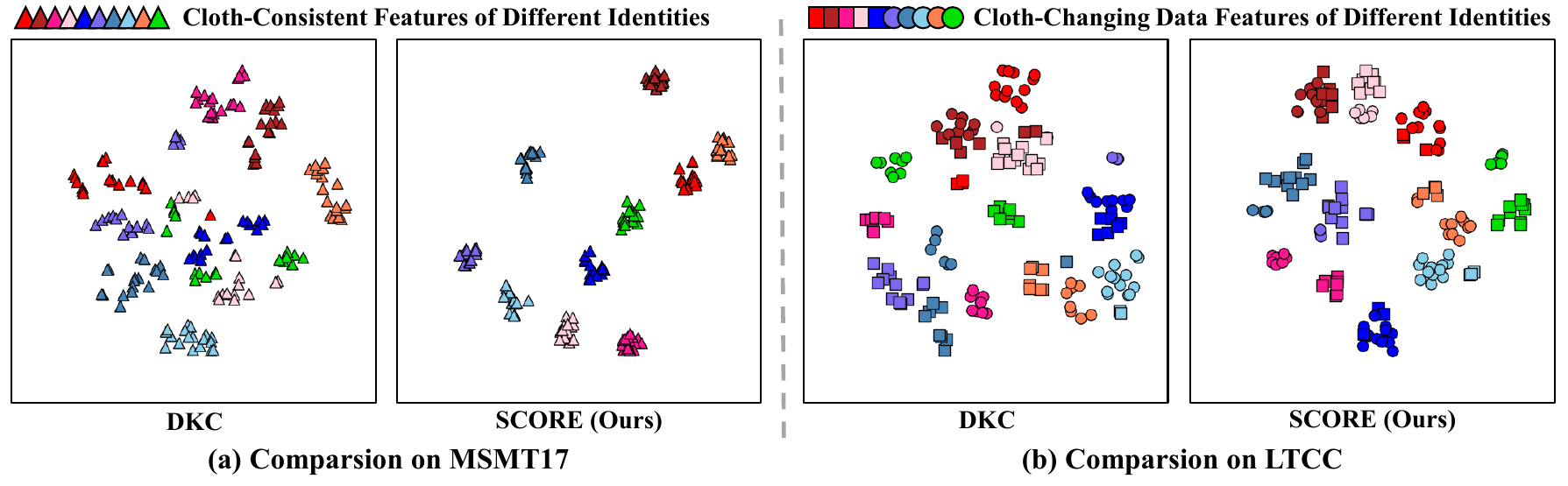}
        \caption{\label{fig:train_features} The t-SNE visualization of cloth-consistent and cloth-changing training data. }
\end{figure}

\subsection{Visualization Results}
We visualize the features of Cloth-Consistent data (MSMT17) and Cloth-Changing data (LTCC) in comparison with the state-of-the-art DKC method, as shown in Figure.~\ref{fig:train_features}. The results indicate that our approach achieves better feature separation across identities on both types of data. The results arise because DKC adopts unlearnable distributions, suffering from limited knowledge representation capacity. In contrast, our SCORE model introduces an adaptive subdistribution learning and reinforcement paradigm, effectively enhancing the consolidation of Cloth-Consistent and Cloth-Changing knowledge.


\section{Conclusion}

In this paper, we focus on the challenging Cloth-Hybrid Lifelong Person Re-Identification (CH-LReID) task and introduce a  novel SubDistribution-aware COllaborative Knowledge REinforcing (SCORE) approach. To address the deteriorated forgetting caused by the cloth-consistent and cloth-changing knowledge conflict, we propose to explicitly model the intra-identity diversity to improve the compatibility of both knowledge. To achieve this, an Adaptive SubDistribution Modeling mechanism is developed, where a set of distributional subprototypes is assigned to each identity to capture the intra-identity diversity. Besides, a  Distributional Knowledge Reinforcement scheme is introduced, where the knowledge of old distributional subprototypes is retained in the new ones by a collaborative aligning mechanism. Extensive experiments demonstrate that our SCORE achieves the state-of-the-art performance. Our approach underscores the potential of automatic subdistribution modeling in improving conflict knowledge integration.



%
%

 \textbf{Acknowledgements}:
This work was a research achievement of National Engineering Research Center of New Electronic Publishing Technologies, and was supported by the National Natural Science Foundation of China (62376011, 62406313), the National Key R\&D Program of China (2024YFA1410000), and the Postdoctoral Fellowship Program of China Postdoctoral Science Foundation, Grant No. YJB20250283.

\bibliographystyle{splncs04}
\bibliography{main}



\title{Supplemental Materials} 

\author{Kunlun Xu\inst{1}\orcidlink{0000-0002-1706-4102} \and
Liangyu Ma\inst{1}\orcidlink{0009-0006-3423-1933} \and
 Jiangmeng Li\inst{2}\orcidlink{0000-0002-3376-1522}
 \and
Xin Tong\inst{3}\orcidlink{0000-0001-7424-0726}
 \and
 Xiaode Liu\inst{3}\orcidlink{0000-0003-3067-4543}
 \and
Yufei Guo\inst{3*}\orcidlink{0000-0002-4920-0965}
 \and
 Jiahuan Zhou\inst{1*}\orcidlink{0000-0002-3301-747X}
 }

\authorrunning{Kunlun Xu et al.}


\institute{Wangxuan Institute of Computer Technology, Peking University, Beijing, China 
\email{\{xkl,2300012949\}@stu.pku.edu.cn},
\email{jiahuanzhou@pku.edu.cn}
\and
University of Chinese Academy of Sciences, Beijing, China
\email{jiangmeng2019@iscas.ac.cn}
\and
Intelligent Science \& Technology Academy of CASIC, Beijing 100041, China\\
\email{\{xin\_tong,lxde,yfguo\}@pku.edu.cn}
}

\maketitle

In the supplementary materials, we provide additional experimental results and detailed analyses to further support our main findings. Specifically, the supplementary materials include:
\begin{itemize}
\item More comprehensive comparisons of SCORE on the Cloth-Consistent LReID benchmark.

\item  Additional experimental comparison with the integration of Cloth-Changing ReID with LReID methods.

\item Details of the relevant ReID datasets and their organization within the CH-LReID benchmark.

\item Analysis of the parameter $k$ which represents the subdistribution number.

\item Analysis of computing and storage costs

\item Analysis of forgetting measure and backward transfer.

\item Additional qualitative results on feature distributions to illustrate the effectiveness of our method.
\end{itemize}

\begin{table*}[htbp]
   \centering
   \caption{Performance comparison on training order-1: Market-1501$\rightarrow$ CUHK-SYSU$\rightarrow$  DukeMTMC$\rightarrow$  MSMT17$\rightarrow$  CUHK03. }
  \scalebox{0.7}{
    \begin{tabular}{llccccccccccc>{\columncolor{seen_back}}c>{\columncolor{seen_back}}c>{\columncolor{unseen_back}}c>{\columncolor{unseen_back}}c}   
    \hline
   &\multirow{2}[1]{*}{Method}&\multirow{2}[1]{*}{Publication} & \multicolumn{2}{c}{Market-1501} & \multicolumn{2}{c}{CUHK-SYSU} & \multicolumn{2}{c}{DukeMTMC} & \multicolumn{2}{c}{MSMT17} & \multicolumn{2}{c}{CUHK03} & \multicolumn{2}{>{\columncolor{seen_back}}c}{\textbf{Seen-Avg}} & \multicolumn{2}{>{\columncolor{unseen_back}}c}{\textbf{Unseen-Avg}} \\
\cline{4-17}          &&& mAP   & R@1   & mAP   & R@1   & mAP   & R@1   & mAP   & R@1   & mAP   & R@1   & mAP   & R@1   & mAP   & R@1 \\
\hline

&Joint-Train&-&75.3&90.1&84.5&86.0&66.9&81.6&31.6&57.1&58.5&61.4&63.4&75.2&55.2&48.2\\

\hline
\multirow{5}[0]{*}{\rotatebox{90}{CIL} } 
&LwF~\cite{li2017learning} &T-PAMI 2017     & 56.3  & 77.1  & 72.9  & 75.1  & 29.6  & 46.5  & 6.0   & 16.6  & 36.1  & 37.5  & 40.2  & 50.6  & 47.2  & 42.6  \\ 

&SPD~\cite{tung2019similarity} &ICCV 2019   & 35.6  & 61.2  & 61.7  & 64.0  & 27.5  & 47.1  & 5.2   & 15.5  & 42.2  &44.3  & 34.4  & 46.4  & 40.4  & 36.6  \\  

&PRD~\cite{asadi2023prototype} &ICML 2023   & 7.3   & 18.0  & 33.5  & 35.6  & 3.7   & 7.6   & 0.8   & 2.4   & 33.8  & 33.8  & 15.8  & 19.5  & 23.0  & 17.7  \\

&PRAKA~\cite{shi2023prototype}& ICCV 2023   & 37.4  & 61.3  & 69.3  & 71.8  & 35.4  & 55.0  & 10.7  & 27.2  & \textbf{54.0}  & \textbf{55.6} & 41.3  & 54.2  & 47.7  & 41.6  \\

&FCS~\cite{li2024fcs} &CVPR 2024            &58.3  & 78.5  & 75.1  & 76.2  & 42.6  & 59.8  & 10.2  & 24.3  & 35.3  & 34.9  & 44.3  & 54.7  & 52.1  & 44.2  \\

\hline
\multirow{9}[0]{*}{\rotatebox{90}{LReID} } 
&CRL~\cite{zhao2021continual} &WACV 2021    & 58.0  & 78.2  & 72.5  & 75.1  & 28.3  & 45.2  & 6.0   & 15.8  & 37.4  & 39.8  & 40.5  & 50.8  & 47.8  & 43.5  \\

&AKA~\cite{pu2021lifelong}&CVPR 2021 & 58.1  & 77.4  & 72.5  & 74.8  & 28.7  & 45.2  & 6.1   & 16.2  & 38.7  & 40.4  & 40.8  & 50.8  & 47.6  & 42.6  \\ 

&PatchKD~\cite{sun2022patch}  & ACM MM 2022  &\underline{68.5}  & \underline{85.7} & 75.6  & 78.6  & 33.8  & 50.4  & 6.5   & 17.0  & 34.1  & 36.8  &43.7  &53.7  &49.1  &45.4  \\

&MEGE~\cite{pu2023memorizing} &T-PAMI 2023  & 39.0    & 61.6  & 73.3  & 76.6  & 16.9  & 30.3  & 4.6   & 13.4  & 36.4  & 37.1  & 34.0    & 43.8  & 47.7  & 44.0 \\

&LSTKC~\cite{xu2024lstkc} & AAAI 2024    &54.7  & 76.0  &81.1  & {83.4 } & 49.4  &66.2  &  \underline{20.0}  &  \underline{43.2}  &\underline{44.7}  &\underline{46.5}  &50.0  &63.1  &57.0  &49.9  \\

&C$^2$R~\cite{cui2024continual} & CVPR 2024    & \textbf{69.0} &   \textbf{86.8} &76.7 & 79.5 & 33.2 & 48.6 & 6.6 & 17.4 & 35.6 &36.2 &44.2 &53.7 & 48.7 & 41.9 \\
&DKP~\cite{xu2024distribution} &CVPR 2024    &60.3	&80.6 &  \textbf{83.6} &  \textbf{85.4} &  \underline{51.6} &  \underline{68.4} & {19.7} & {41.8} &43.6 &44.2 &  \underline{51.8} &  \underline{64.1} &  \underline{59.2} & \underline{51.6}	\\
\hline
&\textbf{SCORE} &This Paper & 65.7	&85.1	&\underline{81.3}	&\underline{83.2}	&\textbf{55.2}	&\textbf{72.5}	&\textbf{24.7}	&\textbf{51.1}	&44.3	&44.1	&\textbf{54.3}	&\textbf{67.2}&   \textbf{63.4}&	\textbf{56.5}\\

\hline

    \end{tabular}%
   }
 \raggedright  \\  
\label{tab:lreid1}%
\end{table*}%

\begin{table*}[htbp]
  \centering
    \caption{Performance comparison on training order-2: DukeMTMC$\rightarrow$ MSMT17$\rightarrow$Market-1501$\rightarrow$ CUHK-SYSU$\rightarrow$ CUHK03. }
  \scalebox{0.7}{
    \begin{tabular}{llccccccccccc>{\columncolor{seen_back}}c>{\columncolor{seen_back}}c>{\columncolor{unseen_back}}c>{\columncolor{unseen_back}}c}  
    \hline
    &\multirow{2}[1]{*}{Method}&\multirow{2}[1]{*}{Publication}& \multicolumn{2}{c}{DukeMTMC} & \multicolumn{2}{c}{MSMT17} & \multicolumn{2}{c}{Market-1501} & \multicolumn{2}{c}{CUHK-SYSU} & \multicolumn{2}{c}{CUHK03} &  \multicolumn{2}{>{\columncolor{seen_back}}c}{\textbf{Seen-Avg}}      & \multicolumn{2}{>{\columncolor{unseen_back}}c}{\textbf{Unseen-Avg}} \\
\cline{4-17}         && & mAP   & R@1   & mAP   & R@1   & mAP   & R@1   & mAP   & R@1   & mAP   & R@1   & mAP   & R@1   & mAP   & R@1 \\
\hline 
 
   &Joint-Train&-&66.9&81.6&31.6&57.1&75.3&90.1&84.5&86.0&58.5&61.4&63.4&75.2&55.2&48.2\\
    \hline
\multirow{5}[0]{*}{\rotatebox{90}{CIL} }   
    &LwF~\cite{li2017learning} &T-PAMI 2017      & 42.7  & 61.7  & 5.1   & 14.3  & 34.4  & 58.6  & 69.9  & 73.0  & 34.1  & 34.1  & 37.2  & 48.4  & 44.0  & 40.1  \\
    &SPD~\cite{tung2019similarity}& ICCV 2019  & 28.5  & 48.5  & 3.7   & 11.5  & 32.3  & 57.4  & 62.1  & 65.0  &43.0  & 45.2  & 33.9  & 45.5  & 39.8  & 36.3  \\
    &PRD~\cite{asadi2023prototype} & ICML 2023  & 3.6   & 8.2   & 0.6   & 1.8   & 8.9   & 22.3  & 34.6  & 36.1  & 35.4  & 35.3  & 16.6  & 20.7  & 20.7  & 15.0  \\
    &PRAKA~\cite{shi2023prototype} & ICCV 2023  & 31.2  & 48.7  &6.6   &19.1  &47.8  &69.8  & 70.4  & 73.0  &  \textbf{54.9}  &  \textbf{56.6}  & 42.2  & 53.4  & 48.4  & 41.1  \\    
    &FCS~\cite{li2024fcs}    &  CVPR 2024   &53.6  & 70.0  & 9.5   & 23.5  & 48.7  & 69.9  & 76.2  & 78.2  & 37.1  & 38.4  & 45.0  & 56.0  & 52.7  & 45.1  \\
    \hline
\multirow{5}[0]{*}{\rotatebox{90}{LReID} } 
    &CRL~\cite{zhao2021continual} &  WACV 2021    & 43.5  & 63.1  & 4.8   & 13.7  & 35.0  & 59.8  & 70.0  & 72.8  & 34.5  & 36.8  & 37.6  & 49.2  & 45.3  & 41.4  \\
    &AKA~\cite{pu2021lifelong} & CVPR 2021   & 42.2  & 60.1  & 5.4   & 15.1  & 37.2  & 59.8  & 71.2  & 73.9  & 36.9  & 37.9  & 38.6  & 49.4  & 46.0  & 41.7  \\
    &PatchKD~\cite{sun2022patch}  & ACM MM 2022         & {58.3} &  {74.1}  & 6.4   & 17.4  & 43.2  & 67.4  &74.5  &76.9  & 33.7  & 34.8  &43.2  &54.1  &48.6  &44.1  \\  

    &MEGE~\cite{pu2023memorizing}& T-PAMI 2023  & 21.6  & 35.5  & 3.0     & 9.3   & 25.0    & 49.8  & 69.9  & 73.1  & 34.7  & 35.1  & 30.8  & 40.6  & 44.3  & 41.1 \\ 
    &LSTKC~\cite{xu2024lstkc}& AAAI 2024 & 49.9  & 67.6  & \underline{14.6}  & \underline{34.0}  &55.1  &76.7  &\underline{82.3}  &\underline{83.8}  &46.3  &48.1  &49.6  &62.1  &57.6  &{49.6}  \\
    &C$^2$R~\cite{cui2024continual} & CVPR 2024    & \underline{59.7} &  \underline{75.0} &7.3 & 19.2 & 42.4 & 66.5 & 76.0 & 77.8 & 37.8 &39.3 &44.7 &55.6 & 48.5 & 41.7  \\
    &DKP~\cite{xu2024distribution} &CVPR 2024 &53.4	&70.5	&{14.5}	&{33.3}	&\underline{60.6}	& \underline{81.0}	&\textbf{83.0}	&\textbf{84.9}	&45.0	&46.1	& \underline{51.3}	& \underline{63.2}   & \underline{59.0}	&\underline{51.6}\\
\hline

&\textbf{SCORE} &This Paper &\textbf{59.8}	&\textbf{76.0}	&\textbf{22.4}	&\textbf{48.3}	&\textbf{63.8}&	\textbf{82.8}	&82.0	&83.1	&44.8	&46.0	&\textbf{54.6}	&\textbf{67.3} &\textbf{62.2}	&\textbf{54.4}\\
\hline
    
    \end{tabular}%
   }
 \raggedright

  \label{tab:lreid2}%
\end{table*}%
\section{Experimental Results on Cloth-Consistent LReID Benchmarks}
We follow the experimental setup of previous Cloth-Consistent LReID works~\cite{xu2024distribution,xu2024lstkc,cui2024continual}, adopting two training orders: Training Order-1: Market-1501$\rightarrow$ CUHK-SYSU$\rightarrow$ DukeMTMC$\rightarrow$ MSMT17$\rightarrow$ CUHK03, and Training Order-2: DukeMTMC$\rightarrow$ MSMT17$\rightarrow$ Market-1501$\rightarrow$ CUHK-SYSU$\rightarrow$ CUHK03. As shown in Table.~\ref{tab:lreid1} and Table.~\ref{tab:lreid2}, our proposed SCORE method achieves competitive performance compared to existing approaches.
Specifically, SCORE achieves an average mAP/R@1 of \textbf{54.3\%/67.2\%} on the seen datasets under training Order-1, and \textbf{54.6\%/67.3\%} under training Order-2. These consistent results across different training sequences demonstrate the robustness of SCORE in retaining previously learned identity-discriminative information. The performance indicates that SCORE not only captures fine-grained intra-identity variations but also facilitates collaborative knowledge retention at both the distributional and instance levels. By aligning new and historical subdistributions while preserving instance-level structural relations, SCORE effectively mitigates the effects of catastrophic forgetting and maintains identity consistency throughout continual learning stages.


In addition, on the unseen datasets from training Order-1 and Order-2, SCORE achieves an average mAP/R@1 of \textbf{63.4\%/56.5\%} and \textbf{62.2\%/54.4\%}, respectively. These results highlight the strong generalization ability of our method when applied to previously unseen domains. Such performance gains can be attributed to SCORE’s capacity to model intra-identity variations via learnable distributional subprototypes. By explicitly capturing the underlying subdistributions within each identity, SCORE enables the model to leverage diverse and complementary knowledge representations, rather than overfitting to appearance-specific cues. This facilitates fine-grained person discrimination even under distribution shifts and novel clothing conditions, making the model more robust in real-world, open-set deployment scenarios.

\section{Additional Comparison with Integrating Cloth-Changing ReID with LReID Methods}
To empirically validate whether simply combining state-of-the-art cloth-changing ReID techniques with lifelong ReID (LReID) frameworks can effectively mitigate knowledge conflicts in hybrid streams, we integrated the leading LReID method, DASK, with two representative cloth-changing approaches: DLCR~\cite{dlcr} and PFAC~\cite{pfac}. Both DLCR and PFAC focus on learning clothing-irrelevant features primarily through data augmentation. As detailed in Table~\ref{tab:combined}, our proposed SCORE method significantly outperforms these hybrid baselines. Specifically, SCORE achieves average mAP/R@1 improvements of \textbf{12.4\%/9.6\%} over DASK+DLCR and \textbf{7.5\%/4.3\%} over DASK+PFAC. These substantial gains underscore that SCORE's adaptive subdistribution modeling not only captures intra-identity diversity more effectively but also successfully reconciles the conflicting objectives of cloth-consistent and cloth-changing knowledge, thereby delivering superior compatibility and overall performance.

\begin{table}[t]
  \centering  
  \caption{Comparison with the integration of cloth-changing and LReID methods.}
  \scalebox{0.93}{
    \begin{tabular}{ccc>{\columncolor{gray!20}}c}
    \hline
    Method              & DASK+DLCR   &DASK+PFAC   &\textbf{Ours}   \\
    \hline
     Average mAP/R@1   & 35.4/50.1            &   40.3/55.4        &\textbf{47.8}/\textbf{59.7}      \\
     \hline
    \end{tabular}
    }      
    \label{tab:combined}%
\end{table}

\section{Analysis of Parameter $k$}
To address concerns regarding the fixed number of subprototypes per identity, we conducted an empirical investigation into a dynamic variant that automatically adjusts the number of subdistributions by adding or removing them upon detecting feature outliers. However, as reported in Table~\ref{tab:dynamic k}, this dynamic $k$ strategy leads to a slight performance degradation of \textbf{0.6\%/1.0\%} in average mAP/R@1 compared to our fixed-$k$ setting. 

We attribute this decline to optimization instability induced by frequent structural changes in the subdistribution space during training. In contrast, fixing $k$ (set to 5) achieves a superior trade-off between modeling flexibility and training stability. These results empirically validate the rationality of our design choice.

\begin{table}[ht]
  \centering  
  \caption{Comparison between dynamic and fixed number of subprototypes.}
  \scalebox{0.93}{
    \begin{tabular}{cc>{\columncolor{gray!20}}cc}
    \hline
    Method              & dynamic $k$   &\textbf{fixed $k$ (Ours)}   \\
    \hline
     Average mAP/R@1   & 47.2/58.7               &\textbf{47.8}/\textbf{59.7}      \\
     \hline
    \end{tabular}
    }    
    \label{tab:dynamic k}%
\end{table}

\section{Analysis of Forgetting Measure and Backward Transfer}

To further quantify forgetting behaviors using direct metrics, we evaluated Backward Transfer (BWT) and Forgetting Measure (FM) on both mAP and R@1, benchmarking our approach against the recent state-of-the-art method, DKC~\cite{cui2025dkc}. The results demonstrate that SCORE significantly surpasses DKC, achieving improvements of \textbf{4.8/2.3} in BWT and \textbf{5.6/3.2} in FM (mAP/R@1). These higher scores indicate substantially reduced negative backward transfer and minimized forgetting of previously acquired knowledge. We attribute these gains to SCORE's adaptive subdistribution modeling mechanism, which explicitly mitigates inter-domain knowledge conflicts by preserving distinct appearance modes for each identity and reinforcing distributional consistency across learning stages. The consistent superiority across both metrics confirms SCORE's effectiveness in maintaining long-term knowledge retention while seamlessly adapting to new domains.

\begin{table*}[ht]
  \centering  
  \caption{Backward Transfer (BWT) and Forgetting Measure (FM) metrics in comparison with SOTA DKC.} 
  \scalebox{0.93}{
    \begin{tabular}{cc>{\columncolor{gray!20}}cc}
    \hline
    Method              & DKC   &\textbf{Ours}   \\
    \hline
     Average BWT(mAP)/BWT(R@1)   &   -8.2/-4.0         &\textbf{-3.4}/\textbf{-1.7}      \\
     \hline
     Average FM(mAP)/FM(R@1)   &   9.1/5.4         &\textbf{3.5}/\textbf{2.2}      \\
     \hline
    \end{tabular}
    }  
    \label{tab:forget}%
\end{table*}

\section{Anaylsis of Computing and Storage Costs}

Although our method involves maintaining multiple sub-distribution parameters for each identity—along with operations such as stochastic sampling, shift matrix calculation, and relational matrix distillation—we conducted a comprehensive evaluation of its computational and storage overhead. As demonstrated in Table~\ref{tab:costs}, the sub-distribution parameters and their associated computations account for merely \textbf{4.7\%} and \textbf{6.0\%} of the total overhead, respectively. This marginal increase in GPU memory usage is attributed to the fact that sub-distribution modeling relies primarily on lightweight matrix operations that are decoupled from the heavy backbone network, thereby avoiding significant additional memory consumption.

\begin{table}[ht]
  \centering  
  \caption{Computing and storage costs.}
  \scalebox{0.93}{
    \begin{tabular}{cccc}
    \hline
                  & GPU memory   &storage  \\
    \hline
   \rowcolor{gray!20}   Subdistribution/Total  & 832.0MB/17892.3MB (4.7\%)   &6.0MB/100.2MB (6.0\%)    \\
     \hline
    \end{tabular}
    }  
    \label{tab:costs}%
\end{table}

\section{Dataset Details}
The CH-LReID benchmark~\cite{cui2025dkc} is constructed upon existing LReID and CC-ReID settings and consists of seven datasets. It incorporates three widely used cloth-consistent datasets (Market-1501~\cite{zheng2015scalable}, MSMT17-V2 (MSMT17)~\cite{wei2018person}, and CUHK03~\cite{li2014deepreid}) as well as two representative cloth-changing datasets (LTCC~\cite{qian2020long} and PRCC~\cite{yang2019person}).

To alleviate data imbalance across datasets while maintaining sufficient identity diversity, 500 identities are selected from each cloth-consistent dataset, whereas all training identities are retained for cloth-changing datasets. This configuration forms the standardized CH-LReID benchmark. As summarized in Table~\ref{tab:datasets}, the column “Original Identities” denotes the total number of identities in each dataset, while “CH-LReID Identities” indicates the subset of identities included in our benchmark.

Furthermore, it is worth noting that our proposed method does not introduce ethical or regulatory concerns, because all experiments are conducted on the CH-LReID benchmark proposed in a prior CVPR 2024 study~\cite{cui2025dkc}, and the datasets in the CH-LReID benchmark are publicly available and have been widely used in major conferences. Therefore, the data usage in this work complies with standard research norms.

\begin{table*}[t]
  \centering  
  \caption{The statistics of datasets used in the CH-LReID benchmark.} 
  \scalebox{0.93}{
    \begin{tabular}{l|l|llc|ccc}
    \hline
    \multirow{2}[0]{*}{Type} & \multirow{2}[0]{*}{Datasets Name} & \multicolumn{3}{c|}{Original Identities} & \multicolumn{3}{c}{CH-LReID Identities} \\
\cline{3-8}          &       & Train & Query & Gallery & Train & Query & Gallery \\
    \hline
    \multirow{3}[2]{*}{Cloth-Consistent} 
          & Market-1501~\cite{zheng2015scalable} & 751   & 750   & 751   & 500   & 751   & 751 \\
          & MSMT17-V2~\cite{wei2018person}& 1041  & 3060  & 3060  & 500   & 3060  & 3060 \\  
          & CUHK03~\cite{li2014deepreid} & 767   & 700   & 700   & 500   & 700   & 700 \\
          
    \hline
    \multirow{2}[2]{*}{Cloth-Changing}
    & LTCC~\cite{qian2020long}& 77   & 75    & 75    & 77   & 75    & 75 \\
    & PRCC~\cite{yang2019person} & 150   & 71   & 71   & 150   & 71   & 71 \\

    \hline
    \end{tabular}%
    }
      
    \label{tab:datasets}%
\end{table*}%

\begin{figure*}[t]
    \centering
	\includegraphics[width=0.8\linewidth]{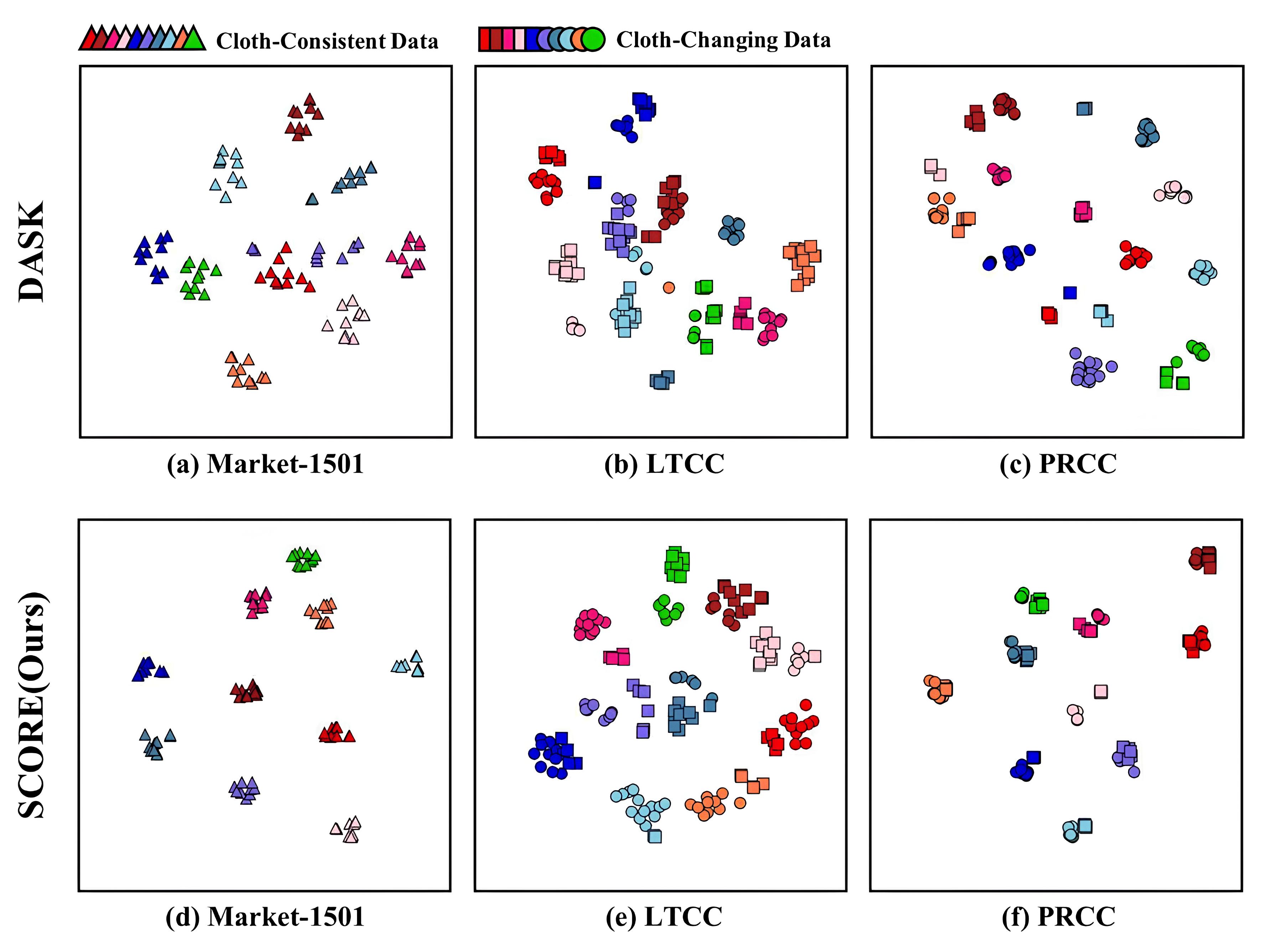}
        \caption{\label{fig:visual1} The t-SNE visualization of cloth-consistent and cloth-changing training data. }
\end{figure*}

\section{More Visualization Results on Features}

We visualize the features of different identities derived from both cloth-changing and cloth-consistent training data, as shown in Fig.~\ref{fig:visual1}. It is clearly observed that our SCORE method performs consistent learning across both cloth-consistent and cloth-changing data in the CH-LReID scenario. By collaboratively leveraging differentiated knowledge, SCORE effectively alleviates knowledge conflicts and mitigates catastrophic forgetting.



%
%

\end{document}